\documentclass[accepted]{uai2023} 
\usepackage[british]{babel}

\usepackage{natbib}
\usepackage{mathtools} 
\usepackage{booktabs}
\usepackage[capitalize]{cleveref}
\usepackage{algorithm}
\usepackage{algorithmic}
\usepackage{xcolor}
\usepackage{xspace}
\usepackage{ifthen}

\newcommand{\includeappendix}{yes}

\newcommand{\appendixDenseDeriv}{%
    \ifthenelse{\equal{\includeappendix}{yes}}%
      {Appendix~\ref{app:densities_deriv}}%
      {Appendix A}\xspace
}
\newcommand{\appendixDerivDense}{%
    \ifthenelse{\equal{\includeappendix}{yes}}%
      {Appendix~\ref{appendix:deriving_densities}}%
      {Appendix A.7}\xspace
}
\DeclareRobustCommand{\appendixDetailedResults}{%
    \ifthenelse{\equal{\includeappendix}{yes}}%
      {Appendix~\ref{appendix:detailed_results}}%
      {Appendix B}\xspace
}

\usepackage{amsthm, amsmath, amssymb} % Mathematical typesetting

\renewcommand{\v}[1]{\boldsymbol{\mathbf{#1}}}

\newcommand{\bcket}[3]{\left#1 #3 \right#2}

\renewcommand{\b}{\bcket{(}{)}}

\newcommand{\sqb}{\bcket{[}{]}}
\newcommand{\abs}{\bcket{\lvert}{\rvert}}
\newcommand{\cb}{\bcket{\{}{\}}}

\newcommand{\K}{\mathbf{K}}

\newcommand{\GW}{\mathcal{GW}}
\newcommand{\AGW}{\mathrm{A}\text{-}\mathcal{GW}}
\newcommand{\ABGW}{\mathrm{AB}\text{-}\mathcal{GW}}
\newcommand{\G}{\mathbf{G}}
\newcommand{\V}{\mathbf{V}}
\newcommand{\R}{\mathbb{R}}
\newcommand{\Rm}{\mathbf{R}}
\renewcommand{\L}{\mathbf{L}}

\newcommand{\wM}{\mathbf W} %Wishart Matrix
\newcommand{\nD}{P} %Number of datapoints
\newcommand{\E}{\mathbb{E}}

\renewcommand{\P}[1][]{\operatorname{P}_{#1}\b}

\newcommand{\qd}[1][]{\operatorname{Q}_{{#1}}\b}
\newcommand{\pd}[1][]{\operatorname{P}_{{#1}}\b}
\newcommand{\pt}{\operatorname{P}}

\newcommand{\chol}{\mathrm{chol}}
\newcommand{\normal}{\mathcal{N}\b}
\newcommand{\gammad}{\operatorname{Gamma}\b}
\newcommand{\gammat}{\operatorname{Gamma}}
\newcommand{\siid}{\sim_{\mathrm{iid}}}

\newcommand{\QGW}{\operatorname{Q}_{\GW}}
\newcommand{\QAGW}{\operatorname{Q}_{\AGW}}
\newcommand{\QABGW}{\operatorname{Q}_{\ABGW}}

\newcommand{\boston}{\textsc{Boston}}
\newcommand{\concrete}{\textsc{Concrete}}
\newcommand{\energy}{\textsc{Energy}}
\newcommand{\kinnm}{\textsc{Kin8nm}}
\newcommand{\naval}{\textsc{Naval}}
\newcommand{\power}{\textsc{Power}}
\newcommand{\protein}{\textsc{Protein}}
\newcommand{\wine}{\textsc{Wine}}
\newcommand{\yacht}{\textsc{Yacht}}

\newcommand{\Wish}[1]{\mathcal{W}\b{#1}}

\newcommand{\F}{\mathbf{F}}
\renewcommand{\H}{\mathbf{H}}
\newcommand{\0}{{\mathbf{0}}}
\newcommand{\N}{\mathcal{N}\b}
\newcommand{\I}{\mathbf{I}}
\newcommand{\transpose}[1]{{#1}^T}
\renewcommand{\L}{\mathbf{L}}
\renewcommand{\S}{\mathbf{S}}
\newcommand{\reals}{\mathbb{R}}

\DeclareMathOperator*{\etr}{\mathrm{etr}}
\newcommand{\T}{\mathbf{T}}
\newcommand{\bsc}[2]{\left( #1 ; #2 \right)}
\newcommand{\Nc}{\mathcal{N}\bsc}

\newcommand{\A}{\mathbf{A}}

\newcommand{\Y}{\mathbf{Y}}
\newcommand{\X}{\mathbf{X}}
\newcommand{\La}{\mathbf{\Lambda}}
\newcommand{\nt}{{\tilde{\nu}}}
\newcommand{\B}{\mathbf{B}}
\newcommand{\C}{\mathbf{C}}
\newcommand{\D}{\mathbf{D}}
\newcommand{\dd}[2][]{\frac{\partial #1}{\partial #2}}

\renewcommand{\L}{\mathbf{L}}
\newcommand{\Q}[1][]{\operatorname{Q}_{#1}\b}
\newcommand{\MN}{\mathcal{MN}\b}

\title{An Improved Variational Approximate Posterior for the Deep Wishart Process}

\author[1]{Sebastian Ober}
\author[2]{Ben Anson}
\author[2]{Edward Milsom}
\author[2]{Laurence Aitchison}
\affil[1]{%
    University of Cambridge
}
\affil[2]{%
    University of Bristol
}
\begin{document}
\maketitle
\begin{abstract}
  Deep kernel processes are a recently introduced class of deep Bayesian models that have the flexibility of neural networks, but work entirely with Gram matrices. They operate by alternately sampling a Gram matrix from a distribution over positive semi-definite matrices, and applying a deterministic transformation. When the distribution is chosen to be Wishart, the model is called a deep Wishart process (DWP). This particular model is of interest because its prior is equivalent to a deep Gaussian process (DGP) prior, but at the same time it is invariant to rotational symmetries, leading to a simpler posterior distribution. Practical inference in the DWP was made possible in recent work (``A variational approximate posterior for the deep Wishart process" \citealp{ober2021vardwp}) where the authors used a generalisation of the Bartlett decomposition of the Wishart distribution as the variational approximate posterior. However, predictive performance in that paper was less impressive than one might expect, with the DWP only beating a DGP on a few of the UCI datasets used for comparison. In this paper, we show that further generalising their distribution to allow linear combinations of rows and columns in the Bartlett decomposition results in better predictive performance, while incurring negligible additional computation cost.
\end{abstract}
\section{Introduction}\label{sec:intro}
Deep kernel processes (DKPs) \citep{aitchison2021dkp} are a class of deep Bayesian models which have the flexibility of neural networks (NNs), but work entirely with Gram matrices. 
NNs have many tuneable parameters which allow them to automatically adapt to problems, and therefore learn good top-layer representations, which turns out to be very important for complex tasks like image classification \citep{krizhevsky2012cnn}. 
On the other hand, most kernels only have a very small number of tuneable hyperparameters, meaning that the kernel matrices they produce are comparatively rigid, and do not have the ability, that NNs have, to learn flexible top-layer representations. 
DKPs solve this problem by alternately taking the kernel matrix from the previous layer, and sampling from a distribution over positive semi-definite matrices, centred on the previous kernel. 
Since DKPs never sample features (except for the final outputs), they are distinct from e.g.\ deep Gaussian processes \citep{damianou2013deep} (DGPs), which sample features at every layer.

A particular DKP, called the deep Wishart process (DWP), is of particular interest since \cite{aitchison2021dkp} showed that its prior is equivalent to the DGP prior. 
However, they were unable to perform inference in the DWP due to the lack of a sufficiently flexible yet tractable distribution over positive semi-definite matrices to use as an approximate posterior. 
The first solution to this problem was posed by \cite{ober2021vardwp}, who developed a generalisation of the Bartlett decomposition of the Wishart distribution, and used it as the basis of their approximate posterior in a series of experiments that compared DWPs to DGPs. 
In theory, purely kernel-based methods should have an advantage over feature-based methods, since Gram matrices are invariant to certain symmetries to which feature-based methods are not, leading to simpler posteriors (see Appendix D2 in \citealt{aitchison2021dkp}). 
However, the experiments in~\citet{ober2021vardwp} showed only minor advantages over DGPs on a fraction of the datasets they tested. 
In this paper, we extend the generalised (singular) Wishart distributions proposed by \cite{ober2021vardwp} by introducing parameters that rotate, stretch, and mix the rows and columns of the Bartlett decomposition, and show that this added flexibility in the approximate posterior allows DWPs to consistently match or outperform DGPs on UCI datasets, while adding negligible computation cost.
\section{Contributions}
Concretely, our contributions are:
\begin{itemize}
    \item We propose the A-generalised (singular) Wishart and AB-generalised (singular) Wishart distributions, two flexible distributions over positive semi-definite matrices, and we provide full derivations for the densities of these distributions in \appendixDenseDeriv.
    \item We prove both analytically and empirically that the A/AB-generalised (singular) Wishart families are proper supersets of the generalised (singular) Wishart family proposed by \cite{ober2021vardwp}.
    \item We show experimentally that our proposed approximate posteriors provide significant performance benefits on UCI datasets, while adding negligible computation cost.
\end{itemize}
\section{Related Work}
Perhaps the closest prior work is \citet{ober2021vardwp}, which introduces generalised (singular) Wishart approximate posteriors for the deep Wishart process.
However, the performance in that paper was less impressive than one might have expected, indicating that there may be room to further improve the family of approximate posteriors over Gram matrices.
We provide such an improvement by introducing A and AB-generalised (singular) Wishart approximate posteriors, which exhibit considerably improved performance over the original approximate posterior from \citet{ober2021vardwp}.

The deep kernel process line of work emerged from \citet{aitchison2021dkp}.
While they introduced the deep Wishart process prior, they were not able to perform inference, as they did not have a suitable approximate posterior (that approximate posterior was developed in \citealt{ober2021vardwp}).
Instead, they were able to do inference in the alternative deep inverse Wishart process, which (unlike the deep Wishart process) does not have any equivalences to DGPs.

The deep kernel process research direction was originally inspired by work showing that infinite-width Bayesian neural networks have GP-distributed outputs \citep{lee2017deep,matthews2018gaussian,novak2018bayesian,garriga2018deep}.
However, this limit is problematic in that the resulting GP kernel is a fixed, deterministic function of the inputs that cannot be learned from data.
Thus, this limit eliminates representation or feature learning, which is perhaps the key mechanism behind the excellent practical performance of neural networks \citep{yang2020feature,aitchison2020bigger}. 
Deep kernel processes \citep{aitchison2021dkp} were inspired by infinite width NNs but designed specifically to retain flexible, learned kernels.

Another related approach that enables representation learning in infinite-width NNs is the deep kernel machine (DKM)~\citep{dkm23,cdkm23}. DKMs differ from DKPs slightly because they are deterministic and correspond directly to an infinitely wide DGP with an infinitely wide top layer~\citep{dkm23}.
\section{Background}\label{sec:background}
In order to understand deep Wishart processes, it is necessary
to first define the Wishart distribution. Our implementation of deep Wishart processes further requires a flexibile approximate posterior. This motivates our proposed A/AB-generalised (singular) Wishart distributions, which in turn are obtained by considering the Bartlett decomposition.
\subsection{Wishart distribution}
\label{sec:wish}
The Wishart distribution is a generalisation of the gamma distribution to positive semi-definite matrices. Suppose we take a matrix ${\mathbf F \in \mathbb{R}^{\nD \times \nu}}$ whose columns ${\mathbf f_\lambda \in \mathbb R^\nD \sim \mathcal N(\mathbf 0,\mathbf \Sigma)}$ are multivariate Gaussian distributed vectors with $\lambda \in \{1,\dotsc,\nu\}$ and $\mathbf \Sigma \in \mathbb{R}^{\nD\times\nD}$, where $\nD$ is the number of datapoints. Then,
\begin{equation}
    \wM \coloneqq \mathbf F \mathbf F^T = \sum_{\lambda=1}^\nu\mathbf f_\lambda\mathbf f_\lambda^T
\end{equation}
is said to be Wishart distributed, denoted ${\wM \in \mathbb R^{\nD \times \nD} \sim \mathcal W(\mathbf \Sigma, \nu)}$, with positive definite scale matrix $\mathbf \Sigma$ and degrees of freedom $\nu$.
\subsection{Deep Wishart Process}
The deep Wishart process is a specific instantiation of a deep kernel process~\citep{aitchison2021dkp}. Moreover, DGPs can be reframed as deep Wishart processes. Consider a DGP model, which samples features $\mathbf F_\ell \in \mathbb R^{\nD \times \nu_\ell}$ at each layer sequentially from a Gaussian process,
\begin{equation}
\label{eq:F|F}
P(\mathbf F_\ell \mid \mathbf F_{\ell-1}) = \prod_{\lambda=1}^{\nu_\ell} \mathcal N(\mathbf f^\ell_\lambda;\,\mathbf 0, \mathbf K(\mathbf F_{\ell-1})).
\end{equation}
The columns $\mathbf f^\ell_\lambda$ are IID multivariate Gaussian random variables, where we apply a kernel function ${k(\cdot, \cdot):\mathbb R^{\nu_\ell} \times \mathbb R^{\nu_\ell} \to \mathbb R}$ pairwise to the previous layer to form the covariance matrix $\mathbf K(\mathbf F_{\ell-1}) \in \mathbb R^{\nD \times \nD}$.

With deep kernel processes, instead of working with features $\mathbf F_\ell$, we work with Gram matrices $\mathbf G_\ell$,
\begin{equation}
    \mathbf{G_\ell} = \frac{1}{\nu_\ell}\mathbf F_\ell \mathbf F_\ell^T.
\end{equation}
With $\mathbf{F}_\ell$ defined by Eq.~\eqref{eq:F|F}, $\mathbf{G}_\ell$ is sampled using the same generative process that defines the Wishart distribution (Sec.~\ref{sec:wish}), so we have,
\begin{equation}
    \mathbf G_\ell \mid \mathbf F_{\ell-1} \sim \mathcal{W}\b{ \frac{1}{\nu_\ell}\mathbf K(\mathbf F_{\ell-1}), \nu_\ell}.
\end{equation}
The final ingredient for defining a deep Wishart process is the fact that we can often compute the kernel matrix $\mathbf K$ from $\mathbf G_{\ell-1}$, without having to know $\mathbf F_{\ell-1}$. This is true for many common kernels, including isotropic kernels, which only depend on the average squared euclidean distance $R^{\ell-1}_{ij}$ between datapoints,
\begin{subequations}
    \begin{align}
         R_{ij}^{\ell} &= \frac{1}{\nu_{\ell}} \sum_{\lambda=1}^{\nu_{\ell}} (F^\ell_{i\lambda} - F^{\ell}_{j\lambda})^2 \\
         &= \frac{1}{\nu_{\ell}} \sum_{\lambda=1}^{\nu_{\ell}} {(F^{\ell}_{i \lambda}})^2 - 2F^{\ell}_{i \lambda}F^{\ell}_{j \lambda} + \b{F^{\ell}_{j \lambda}}^2 \\
         &= G^{\ell}_{ii} - 2 G^{\ell}_{ij} + G^{\ell}_{jj}.
    \end{align}
\end{subequations}
(see \citealp{aitchison2021dkp} for further details).
Hence we can use $\mathbf K\b{\mathbf G_{\ell -1}}$, eliminating features to work entirely with gram matrices, defining our deep Wishart process like so,
\begin{subequations}
    \begin{align}
        \pd{\mathbf G_\ell \mid \mathbf G_{\ell-1}} &= \mathcal{W} \b{\mathbf G_\ell;\,\frac{1}{\nu_\ell}\mathbf K(\mathbf G_{\ell-1}), \nu_\ell},\\
        \pd{\mathbf F_{L+1} \mid \mathbf{\mathbf G_L}} &= \prod_\lambda^{\nu_{L+1}} \mathcal{N} \b{\mathbf f^{L+1}_\lambda ;\, \mathbf 0, \mathbf K(\mathbf G_L)},\label{eq:dwp_output}
    \end{align}
\end{subequations}
where $L$ is the number of hidden layers, $\mathbf G_0 = \frac{1}{\nu_0} \mathbf X \mathbf X^T$ for input data $\mathbf X \in \mathbb R^{\nD \times \nu_0}$, and at the output layer (Eq. \ref{eq:dwp_output}) we sample features that can be provided to a likelihood function $\pd{\mathbf Y \mid \mathbf F_{L+1}}$, e.g. a Gaussian likelihood for regression, or a categorical likelihood for classification.

\subsection{Variational Inference in DWP\lowercase{s}}
As is the case with almost all Bayesian models of reasonable complexity, the true posterior $\pd{\mathbf G_1, \cdots, \mathbf G_L \mid \mathbf X, \mathbf Y}$ is intractable. 
We therefore use variational inference (VI), which replaces the true posterior with an approximate posterior $\qd{\mathbf G_1, \cdots, \mathbf G_L}$. 
This distribution is taken from a variational family of distributions with parameters $\phi$, which are optimised to maximise a lower bound on the marginal log-likelihood of the data.

We consider approximate posteriors that factorise layerwise,
\begin{equation}
    \qd{\G_1, \cdots, \G_L} = \prod_{\ell=1}^L \qd{\G_\ell \mid \G_{\ell-1}},
\end{equation}
where each term $\qd{\G_\ell \mid \G_{\ell-1}}$ is a distribution over positive definite matrices. Note that although the prior of each layer is Wishart distributed, the posterior is not Wishart distributed in general. The seemingly obvious choice for this variational family is the Wishart family itself, but as \cite{aitchison2021dkp} argued, this is not flexible enough. For $\mathbf G \sim \mathcal W(\mathbf \Sigma , \nu)$ (particularly in the case where $\nu$ is fixed), the mean and variance cannot be independently specified since we have
\begin{subequations}
    \begin{align}
        \mathbb E[\G] &= \nu \mathbf \Sigma,\\
        \mathbb V[G_{ij}] &= \nu(\Sigma_{ij}^2 + \Sigma_{ii}\Sigma{jj}).
    \end{align}
\end{subequations}
The ability to independently specify the variance is critical for an approximate posterior to be able to capture potentially narrow true posteriors, so we need an alternative.~\cite{aitchison2021dkp} also suggested that a non-central Wishart distribution would be flexible enough to use as an approximate posterior, but its density is too expensive to evaluate as part of the training loop. Hence \cite{aitchison2021dkp} ultimately did not perform inference in the DWP, instead opting to change the model. In a subsequent work, \cite{ober2021vardwp} introduced the generalised (singular) Wishart distribution, which finally allowed practical inference in DWPs, and which our work builds upon. In order to define that distribution, we first need to recap the Bartlett decomposition.

\subsection{The Bartlett decomposition}\label{sec:bartlett_decomp}
The Bartlett decomposition \citep{bartlett1933on} is a factorisation for Wishart random variables. Specifically, if ${\wM \sim \mathcal W(\mathbf I, \nu)}$ is a standard Wishart random variable (that is, it has identity scale matrix, $\mathbf \Sigma = \mathbf I$), then we have,
\begin{equation}
    \wM = \mathbf T \mathbf T^T,
\end{equation}
where $\mathbf T$ is lower triangular, with the square of its diagonals Gamma-distributed, and its off-diagonals Gaussian-distributed,
\begin{subequations}\label{eq:barlett_decomp_eq}
  \begin{align}
  \T &= \begin{pmatrix}
    T_{11}    & \dotsm & 0          \\ 
    \vdots    & \ddots & \vdots     \\ 
    T_{\nD 1} & \dotsm & T_{\nD \nD} \\ 
  \end{pmatrix},\\
    \pd{T_{ii}^2} &= \gammad{T_{ii}^2;\,\tfrac{\nu-i+1}{2}, \tfrac{1}{2}},\label{eq:barlett_decomp_eq_on_diag}\\
    \pd{T_{i > j}} &= \normal{T_{i > j};\,0, 1}\label{eq:barlett_decomp_eq_off_diag},
  \end{align}
\end{subequations}
In particular each element of $\T$ is independent.
For Wishart distributions with non-identity scale matrices, we can compute the Cholesky decomposition $\mathbf \Sigma = \mathbf L \mathbf L^T$, so that $\wM = \mathbf L \mathbf T \mathbf T^T \mathbf L^T$ (this follows from the canonical definition of the Wishart using Gaussian vectors). 
\subsection{Generalised (singular) Wishart Distribution}\label{sec:gswd}
As shown by \cite{ober2021vardwp}, a generalisation of the Wishart distribution can be obtained by allowing the Bartlett decomposition to be more flexible (and by allowing singular matrices \citep{srivastava2003singularwishart}, since the Wishart ordinarily only supports positive definite matrices). Namely, we can introduce parameters $\alpha_j,\beta_j,\mu_{ij},\sigma_{ij}$ such that the decomposition $\mathbf T$ has distribution,
\begin{subequations}\label{eq:generalised_bartlett}
\begin{align}
  \T &= \begin{pmatrix}
    T_{11}    & \dotsm & 0          \\ 
    \vdots    & \ddots & \vdots     \\ 
    T_{\nu 1} & \dotsm & T_{\nu \nu} \\ 
    \vdots    & \ddots & \vdots     \\ 
    T_{\nD 1}   & \dotsm & T_{\nD \nu}
  \end{pmatrix},\\
  \qd{T_{ii}^2} &= \gammad{T_{ii}^2;\,\alpha_i, \beta_j},\, i \in\{1,\ldots,\nu\}, \label{eq:generalised_bartlett_diagonal} \\
  \qd{T_{i > j}} &=\normal{T_{i>j};\,\mu_{ij}, \sigma_{ij}^2}
\end{align}
\end{subequations}
where, in the singular case of $\nu < \nD$, $\mathbf T$ is now a (tall) rectangular matrix, with the upper square block being lower triangular. This more flexible distribution defines a standard generalised (singular) Wishart random variable, denoted $\mathbf T \mathbf T^T \sim \GW\b{\I, \nu, \v \alpha, \v \beta, \v \mu, \v \sigma}$. For the more general case of $\mathbf \Sigma \neq \mathbf I$, we compute the cholesky decomposition $\mathbf \Sigma = \mathbf L \mathbf L^T$ and obtain $\wM \sim \GW\b{\v \Sigma, \nu, \v \alpha, \v \beta, \v \mu, \v \sigma}$ as $\wM = \mathbf L \mathbf T \mathbf T^T \mathbf L^T$. 

For the general case $\wM = \mathbf{LT} \mathbf T^T \mathbf L^T$, ~\cite{ober2021vardwp} showed that the density of $\wM$ is
\begin{multline}
\label{eq:gw_pdf}
  \qd{\wM} = \b{\prod_{j=1}^\nD \frac{1}{L_{jj}^{\min(j,\nu)}}} \\ \prod_{j=1}^{\nt} \frac{\gammat\b{T_{jj}^2;\, \alpha_j, \beta_j}}{T_{jj}^{\nD-j} L_{jj}^{\nD-j+1}} \prod_{i=j+1}^\nD \N{T_{ij};\, \mu_{ij}, \sigma_{ij}^2}. 
\end{multline}
\section{Methods}
\subsection{$\AGW$ and $\ABGW$ distributions}\label{sec:a_ab_gswd}
Whilst the generalised (singular) Wishart distribution represented a big step for approximate inference in DWPs, the experimental results in \cite{ober2021vardwp} indicated much room for improvement. Despite the theoretical advantages that DWPs have over DGPs due to their invariance to certain posterior symmetries, the DGP still outperformed the DWP in a few cases. By contrast, the A-generalised / AB-generalised (singular) Wishart distributions we introduce in this paper allow the DWP to match or outperform the DGP on all datasets we tested.

One issue with the generalised (singular) Wishart distribution is that it is unclear how flexible it is with respect to linear transformations. 
Suppose $\wM \sim \GW\b{\v \Sigma, \nu, \v \alpha, \v \beta, \v \mu, \v \sigma}$ and consider the mapping $\wM \mapsto \wM ' = \Rm \wM \Rm^T$, where ${\Rm\in \R^{\nD \times \nD}}$ is some invertible matrix.
With $\wM$ constructed as in Section \ref{sec:gswd}, i.e.\ $\wM = \mathbf{LT} \mathbf T^T \mathbf L^T$, where $\mathbf T \mathbf T^T \sim \GW\b{\I, \nu, \v \alpha, \v \beta, \v \mu, \v \sigma}$ and $\mathbf \Sigma = \mathbf L \mathbf L^T$, we have $\wM' = \Rm \L \T \T^T \L \Rm^T$. However, since ${\Rm \L}$ is not in general lower triangular, there is no obvious form that suggests $\wM'$ is in general distributed as a generalised (singular) Wishart.
To remedy this, we introduce more flexibility into the generalised (singular) Wishart distribution.
In particular, instead of parameterising the distribution in terms of $\v \Sigma$, and multiplying $\T$ by the Cholesky of $\v \Sigma$, we both parameterise the distribution and multiply $\T$ with an arbitrary invertible matrix of parameters ${\A\in\R^{\nD \times \nD}}$. We write
$\wM = \A \T(\A \T)^T\sim \AGW\b{\A, \nu, \v \alpha, \v \beta, \v \mu, \v \sigma}$
and say that $\wM$ is A-generalised (singular) Wishart distributed. 
If $\wM$ is A-generalised (singular) Wishart distributed, it is clear that for any transformation
$\v R$, the associated Gram matrix $\v \wM'$ remains in the
same family of distributions; in particular, $\wM' = \v R \wM \v R^T \sim \AGW\b{\v R \A, \nu, \v \alpha, \v \beta, \v \mu, \v \sigma}$.

We can understand $\A$ in $\wM = \A \T (\A \T)^T$ as mixing the rows of $\T$ by linear combinations.
This mixing means that the elements of $\A \T$ have a more complex dependency structure than the elements of $\T$. It also raises the question of whether we could
introduce a more complex dependency structure still. We propose that this can be done by additionally mixing the columns of $\T$ with a matrix $\B$, via $\T \B$,
suggesting an additional 
generalisation of the generalised (singular)
Wishart distribution. We write ${\wM = \A \T \B (\A \T\B)^T\sim \ABGW\b{\A, \B, \nu, \v \alpha, \v \beta, \v \mu, \v \sigma}}$, where ${\B\in\R^{\nu\times\nu}}$ is lower triangular and invertible, and say that $\wM$ is AB-generalised (singular) Wishart distributed.

To use the $\AGW$ and $\ABGW$ distributions for VI, it is necessary to obtain expressions for their densities. Since this is non-trivial, the derivations are provided in the \appendixDerivDense, and we simply quote the results here. The density for the $\AGW$ distribution is,
\begin{multline}
\label{eq:agw_pdf}
    \AGW\b{\wM;\, \A, \nu,\, \v{\alpha},\, \v{\beta},\, \v{\mu},\, \v{\sigma}} \\
    = \frac{\abs{\wM_{:\nt, :\nt}}^{(\nu - N - 1)/2}}{\abs{\A}^{\nu}\abs{(\C_A)_{:\nt, :\nt}}^{(\nu - N - 1)/2}} 
    \prod_{j=1}^\nt \frac{\gammat\b{T_{jj}^2;\,\alpha_j, \beta_j}}{T_{jj}^{N-j}}\\
    \prod_{i=j+1}^N \Nc{T_{ij}}{\mu_{ij}, \sigma_{ij}^2},
\end{multline}
where $\tilde{\nu} = \min\cb{\nu, N}$, $\C_A = \T \T^T$, and the notation $\v X_{:a,:b}$ means the submatrix of $\v X$ obtained by taking the first $a$ rows and $b$ columns.
The density for the $\ABGW$ distribution is,
\begin{multline}
\label{eq:abgw_pdf}
    \ABGW\b{\wM;\, \A, \B, \nu,\, \v{\alpha},\, \v{\beta},\, \v{\mu},\, \v{\sigma}} \\
    = \frac{\abs{\wM_{:\nt, :\nt}}^{(\nu - N - 1)/2}}{\abs{\A}^{\nu}\abs{(\C_{AB})_{:\nt, :\nt}}^{(\nu - N - 1)/2}}
    \prod_{j=1}^\nt \frac{\gammat\b{T_{jj}^2;\,\alpha_j, \beta_j}}{T_{jj}^{N-j}B_{jj}^{2(N-j+1)}}\\
    \prod_{i=j+1}^N \Nc{T_{ij}}{\mu_{ij}, \sigma_{ij}^2},
\end{multline}
where $\C_{AB} = (\T \B) (\T \B)^T$ is defined for notational convenience.
Notice that the densities are defined in terms of both $\wM$ and  $\T$. Since $\wM = (\A \T \B)(\A \T \B)^T$, we can see that $\T$ can be recovered by first computing $(\T\B)(\T\B)^T = \A^{-1}\wM \A^{-T}$, from which we can compute $\T\B$ as the cholesky decomposition. Thus $\T$ is recovered by simply right-multiplying $\T\B$ by $\B^{-1}$.

\begin{figure}[t]
    \centering
    \includegraphics[width=0.45\textwidth]{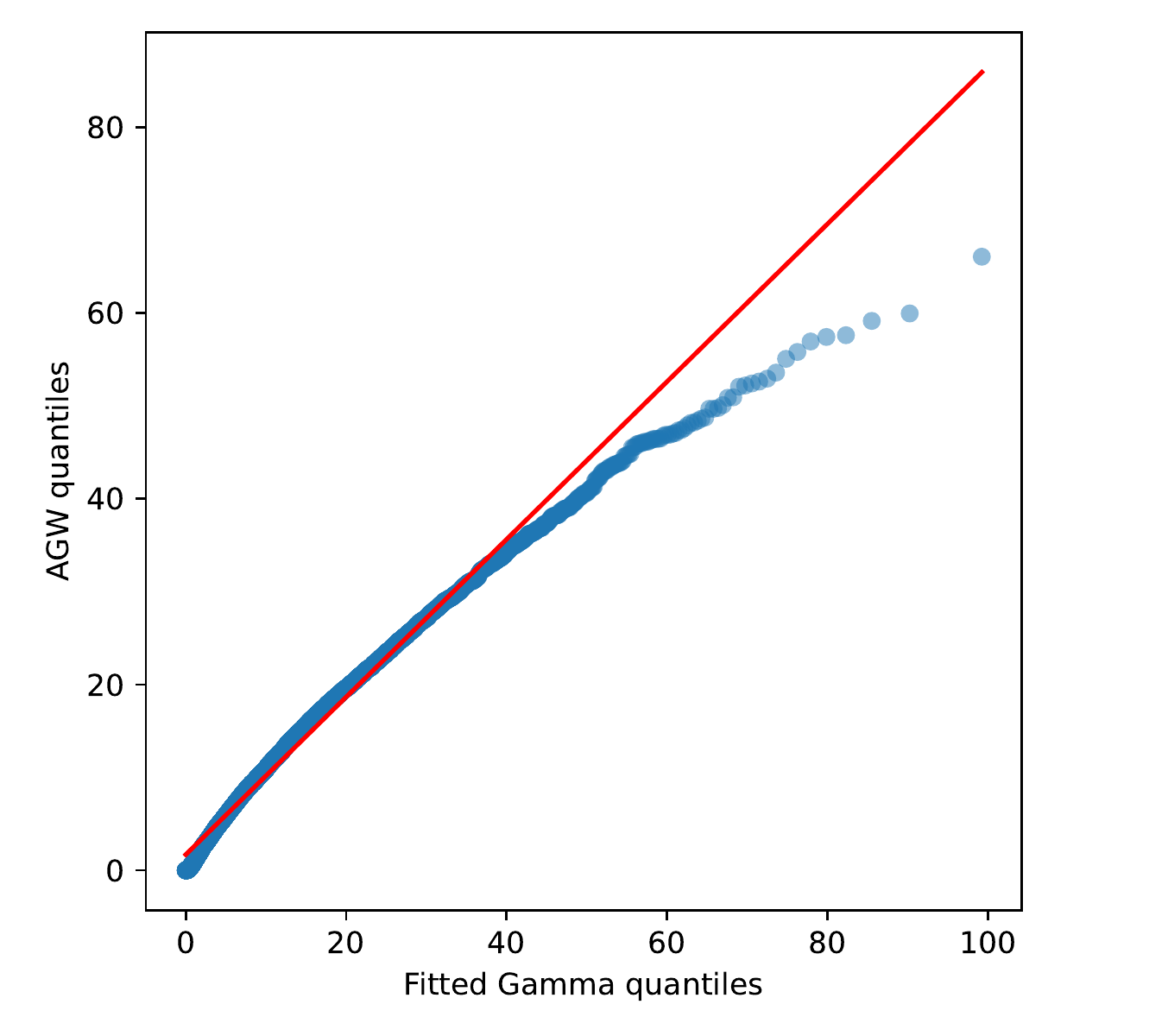}
    \caption{A probability plot comparing the probability density of the top-left element of an $\AGW$ distributed matrix to the density of the top-left element of a $\GW$ distributed matrix. If the two were identically distributed, all the points would lie on the diagonal line.
    We sampled the top-left element of the $\AGW$ distributed matrix 10,000 times using Eq.~\eqref{eq:counter_T}, Eq.~\eqref{eq:counter_W} and Eq.~\eqref{eq:counter_A}.
    We compared this distribution against the closest fitting Gamma distribution, as the top-left element of a $\GW$ distributed matrix is Gamma-distributed (Sec.~\ref{sec:agw_vs_gw}).
    We used $\mu=3, \sigma^2=1$ in Eq.~\eqref{eq:counter_T}.
    In this probability plot, the ``Fitted Gamma quantiles'' are the exact quantiles of the closest fitting Gamma distribution, while the ``AGW quantiles'' are the sample quantiles from the top-left element of the $\AGW$ distribution (i.e.\ the samples ordered by size).
    The plot shows a clear mismatch between the two distributions, confirming the theoretical result that a Gamma distribution does not capture general non-central chi-squared distributions.}
    \label{fig:qq_hist}
\end{figure}
\subsection{$\AGW$ and $\ABGW$ approximate posteriors}
As discussed in Section~\ref{sec:a_ab_gswd}, the A and AB-generalised (singular) Wishart distributions
give us more flexible distributions over Gram matrices, which ought to be useful for VI.
The approximate posterior used by~\cite{ober2021vardwp} was,
\begin{multline}\label{eq:gsw_approx_posterior}
  \qd[\GW]{\G_\ell\mid\G_{\ell-1}}\\
  =
  \GW\big(\G_\ell;\, (1-q_\ell) \tfrac{1}{\nu_\ell} \K\b{\G_{\ell-1}} + q_\ell \V_\ell {\V}_\ell^T,\\ \nu_\ell,\, \v{\alpha}_\ell,\, \v{\beta}_\ell,\, \v{\mu}_\ell,\, \v{\sigma}_\ell\big),
\end{multline}
where $\{\V_\ell, \v{\alpha}_\ell, \v{\beta}_\ell, \v{\mu}_\ell, \v{\sigma}_\ell, q_\ell\}_{\ell=1}^L$ are the
learned variational parameters. Notice that $\V_\ell$ provides flexibility, since $q_\ell$ allows us to control the relative influence of
the kernel from the previous layer, $\K(\G_{\ell-1})$, and an {\it arbitrary, learnable\/} positive (semi-) definite matrix,
$\v V_\ell \v V_\ell^T$.

To use the A and AB generalised (singular) Wishart distributions as approximate posteriors, we obtain $\A_\ell$ by combining an arbitrary invertible matrix of parameters, $\A'_\ell$, with the Cholesky of $(1-q_\ell) \tfrac{1}{\nu_\ell} \K\b{\G_{\ell-1}} + q_\ell \V_\ell {\V}_\ell^T$ to give across-layer dependencies similar to those in the previous $\GW$ approximate posterior (Eq.~\ref{eq:gsw_approx_posterior}),
\begin{align}\label{eq:funny_A_equation}
  \A_\ell = \text{chol}\b{(1-q_\ell) \tfrac{1}{\nu_\ell} \K\b{\G_{\ell-1}} + q_\ell \V_\ell {\V}_\ell^T} \A_\ell'
\end{align}
where $\text{chol}(\cdot)$ returns the lower triangular Cholesky factor.  The $\AGW$ and $\ABGW$ approximate posteriors are then written in terms of this $\A_\ell$:
\begin{subequations}\label{eq:new_approx_post}
\begin{align}
  &\qd[\AGW]{\G_\ell\mid \G_{\ell-1}}\label{eq:agw_approx_post}\\
  &\quad\quad\quad= \AGW\big(\G_\ell;\, \A_\ell,\, \nu_\ell,\, \v{\alpha}_\ell,\, \v{\beta}_\ell,\, \v{\mu}_\ell,\, \v{\sigma}_\ell\big),\nonumber\\
  &\qd[\ABGW]{\G_\ell\mid \G_{\ell-1}}\label{eq:abgw_approx_post}\\
  &\quad\quad\quad= \ABGW\big(\G_\ell;\, \A_\ell,\, \B_\ell,\, \nu_\ell,\, \v{\alpha}_\ell,\, \v{\beta}_\ell,\, \v{\mu}_\ell,\, \v{\sigma}_\ell\big)\nonumber.
\end{align}
\end{subequations}
Here, the variational parameters are $\{\A'_\ell, \V_\ell, \v{\alpha}_\ell, \v{\beta}_\ell, \v{\mu}_\ell, \v{\sigma}_\ell, q_\ell\}_{\ell=1}^L$ for the $\AGW$ approximate posterior, and $\{\A'_\ell, \B_\ell, \v V_\ell, \v{\alpha}_\ell, \v{\beta}_\ell, \v{\mu}_\ell, \v{\sigma}_\ell, q_\ell\}_{\ell=1}^L$ for the $\ABGW$ approximate posterior.

\begin{table*}[ht]
  \caption{
    ELBOs, test log-likelihoods, and test root mean square error for UCI datasets from \citep{pmlr-v48-gal16} for a five-layer network. All metrics are quoted as the mean, plus or minus one standard error, over the splits.
    Better results are highlighted; see \appendixDetailedResults for other depths and additional information.}
  \label{tab:dwp:uci_comb}
  \centering
  \begin{tabular}{lrcccc}
    \toprule
& & & & DWP & \\
& Dataset & DGP & $\QGW$ & $\QAGW$ & $\QABGW$ \\
\midrule 
& \textsc{\textsc{\boston}} & -0.45 $\pm$ 0.00 & -0.37 $\pm$ 0.01 & \textbf{-0.36 $\pm$ 0.00} & \textbf{-0.36 $\pm$ 0.00} \\
& \textsc{\concrete} & -0.50 $\pm$ 0.00 & -0.49 $\pm$ 0.00 & \textbf{-0.45 $\pm$ 0.00} & \textbf{-0.45 $\pm$ 0.00} \\
& \textsc{\energy} & 1.38 $\pm$ 0.00 & 1.40 $\pm$ 0.00 & \textbf{1.42 $\pm$ 0.00} & \textbf{1.41 $\pm$ 0.00} \\
& \textsc{\kinnm} & -0.14 $\pm$ 0.00 & -0.14 $\pm$ 0.00 & \textbf{-0.11 $\pm$ 0.00} & \textbf{-0.11 $\pm$ 0.00} \\
ELBO & \textsc{\naval} & 3.92 $\pm$ 0.04 & 3.59 $\pm$ 0.12 & \textbf{3.97 $\pm$ 0.02} & 3.63 $\pm$ 0.22\\
& \textsc{\power} & \textbf{0.03 $\pm$ 0.00} & 0.02 $\pm$ 0.00 & \textbf{0.03 $\pm$ 0.00} & \textbf{0.03 $\pm$ 0.00} \\
& \textsc{\protein} & \textbf{-1.00 $\pm$ 0.00} & -1.01 $\pm$ 0.00 & \textbf{-1.00 $\pm$ 0.00} & \textbf{-1.00 $\pm$ 0.00}\\
& \textsc{\wine} & -1.19 $\pm$ 0.00 & -1.19 $\pm$ 0.00 & -1.19 $\pm$ 0.00 & -1.19 $\pm$ 0.00 \\
& \textsc{\yacht} & 1.46 $\pm$ 0.02 & 1.59 $\pm$ 0.02 & \textbf{1.79 $\pm$ 0.02} & \textbf{1.79 $\pm$ 0.02} \\
\midrule
& \textsc{\textsc{\boston}} & -2.43 $\pm$ 0.04 & \textbf{-2.38 $\pm$ 0.04}  & -2.39 $\pm$ 0.05 & \textbf{-2.38 $\pm$ 0.04} \\
& \textsc{\concrete} & -3.13 $\pm$ 0.02 & -3.13 $\pm$ 0.02 & \textbf{-3.07 $\pm$ 0.02} & -3.08 $\pm$ 0.02 \\
& \textsc{\energy} & -0.71 $\pm$ 0.03 & -0.71 $\pm$ 0.03 & \textbf{-0.70 $\pm$ 0.03} & \textbf{-0.70 $\pm$ 0.03}\\
& \textsc{\kinnm} & 1.38 $\pm$ 0.00 & 1.40 $\pm$ 0.01 & \textbf{1.41 $\pm$ 0.01} & \textbf{1.41 $\pm$ 0.01} \\
LL & \textsc{\naval} & 8.28 $\pm$ 0.04 & 8.17 $\pm$ 0.07 & \textbf{8.40 $\pm$ 0.02} & 8.10 $\pm$ 0.19 \\
& \textsc{\power} & -2.78 $\pm$ 0.01 & -2.77 $\pm$ 0.01 & \textbf{-2.76 $\pm$ 0.01} & \textbf{-2.76 $\pm$ 0.01} \\
& \textsc{\protein} & -2.73 $\pm$ 0.01 & -2.72 $\pm$ 0.01 & -2.71 $\pm$ 0.01 & \textbf{-2.70 $\pm$ 0.00} \\
& \textsc{\wine} & -0.96 $\pm$ 0.01 & -0.96 $\pm$ 0.01 & -0.96 $\pm$ 0.01 & -0.96 $\pm$ 0.01 \\
& \textsc{\yacht} & -0.73 $\pm$ 0.07 & -0.58 $\pm$ 0.06 & -0.22 $\pm$ 0.09 & \textbf{-0.18 $\pm$ 0.07}\\
\midrule
& \textsc{\textsc{\boston}} & 2.81 $\pm$ 0.14 & 2.82 $\pm$ 0.17 & \textbf{2.77 $\pm$ 0.16} & 2.81 $\pm$ 0.17 \\
& \textsc{\concrete} & 5.49 $\pm$ 0.10 & 5.53 $\pm$ 0.10 & 5.26 $\pm$ 0.11 & \textbf{5.24 $\pm$ 0.11} \\
& \textsc{\energy} & 0.49 $\pm$ 0.01 & \textbf{0.48 $\pm$ 0.01} & \textbf{0.48 $\pm$ 0.01} & \textbf{0.48 $\pm$ 0.01}\\
& \textsc{\kinnm} & 0.06 $\pm$ 0.01 & 0.06 $\pm$ 0.01 & 0.06 $\pm$ 0.00 & 0.06 $\pm$ 0.00 \\
RMSE & \textsc{\naval} & 0.00 $\pm$ 0.00 & 0.00 $\pm$ 0.00 & 0.00 $\pm$ 0.00 & 0.00 $\pm$ 0.00 \\
& \textsc{\power} & 3.88 $\pm$ 0.04 & 3.84 $\pm$ 0.04 & \textbf{3.80 $\pm$ 0.04} & \textbf{3.80 $\pm$ 0.04}\\
& \textsc{\protein} & 3.77 $\pm$ 0.02 & 3.76 $\pm$ 0.02 & 3.73 $\pm$ 0.02 &\textbf{3.70 $\pm$ 0.01} \\
& \textsc{\wine} & 0.63 $\pm$ 0.01 & 0.63 $\pm$ 0.01 & 0.63 $\pm$ 0.01 & 0.63 $\pm$ 0.01 \\
& \textsc{\yacht} & 0.57 $\pm$ 0.05 & 0.50 $\pm$ 0.04 & \textbf{0.37 $\pm$ 0.03} & 0.38 $\pm$ 0.03\\
\bottomrule
 \end{tabular}
\end{table*}

\begin{table*}[t]
  \caption{Average runtime (seconds) for an epoch of \textsc{\boston} and \textsc{\protein}. Error bars were negligible and are excluded.}
  \label{tab:dwp:runtime}
  \centering
  \begin{tabular}{rcccc}
    \toprule
\{dataset\} - \{depth\} & DGP & $\QGW$ & $\QAGW$ & $\QABGW$\\
\midrule 
\textsc{\boston} - 2 & 0.463 & 0.200 & 0.203 & 0.202 \\
5 & 1.292 & 0.358 & 0.373 & 0.370 \\
\midrule 
\textsc{\protein} - 2 & 0.903 & 0.843 & 0.854 & 0.869 \\
5 & 2.012 & 1.806 & 1.846 & 1.839 \\
\bottomrule
  \end{tabular}
\end{table*}
\subsection{$\AGW$ is more flexible than $\GW$}
\label{sec:agw_vs_gw}
To further motivate the utility of the proposed distributions, we now demonstrate that the $\AGW$ family of distributions (Eq.~\ref{eq:agw_pdf}) is a proper superset of the $\GW$ family (Eq.~\ref{eq:gw_pdf}).
The fact that it is a superset can be seen by noting that if we take $\A$ to be lower triangular, and use $\mathbf \Sigma = \A \A^T$, then the $\AGW$ distribution reduces to the $\GW$ distribution.
Note that the $\ABGW$ distribution also reduces to the previous $\GW$ distribution if we take $\A$ to be lower triangular and $\B = \I$. 
We do not make the claim that the $\ABGW$ distribution is strictly more flexible than the $\AGW$ distribution (though it clearly contains the $\AGW$ by just setting $\B = \I$), and instead leave this to future work.

In order to show that the $\AGW$ family contains a proper superset of $\GW$, we must show it contains distributions which the $\GW$ cannot capture. To this end, consider a toy setting where $P=2$ and $\nu=1$, so, $\v T \in \mathbb R^{2 \times 1}$.  We choose this matrix to be,
\begin{equation}
  \label{eq:counter_T}
    \T \coloneqq \begin{pmatrix} g \\ n \end{pmatrix} \sim \begin{pmatrix} \mathcal N\b{0,1} \\ \mathcal N \b{\mu,\sigma^2} \end{pmatrix}
\end{equation}
so that $g^2 \sim \chi^2\b{1} = \gammad{\frac{1}{2} , 2}$. Then we have,
\begin{equation}
  \label{eq:counter_W}
    \wM = \A \T (\A \T)^T \sim \AGW\b{\A, 1, \frac{1}{2}, 2, \mu, \sigma^2}
\end{equation}
where $\wM \in \mathbb R^{2 \times 2}$ and $\A \in \mathbb R^{2 \times 2}$ is any invertible matrix. We shall show that, for certain choices of $\A$, $\wM$ has a distribution that cannot be captured by the $\GW$ distribution. Firstly, we have,
\begin{equation}
    \T \T^T = \begin{pmatrix} g^2 & gn \\ gn & n^2 \end{pmatrix},
\end{equation}
and taking $\v A$ to have concrete value,
\begin{equation}
  \label{eq:counter_A}
    \A = \begin{pmatrix} 1 & 1 \\ 0 & 1 \end{pmatrix},
\end{equation}
we obtain,
\begin{equation}
    \b{\A \T} \b{\A \T}^T = \begin{pmatrix} g^2 + 2gn + n^2 & gn + n^2 \\ gn + n^2 & n^2\end{pmatrix}.
\end{equation}
We need only consider the distribution of the top-left element. We have $g^2 + 2gn + n^2 = (g+n)^2 = X^2$ where $X \sim \mathcal N\b{\mu,\sigma^2 + 1}$, since $g$ and $n$ are normally distributed. 
Hence if we choose $\mu\neq 0$, the top-left element, $X^2$, is noncentral chi-squared distributed.

To conclude the proof, we show that the top-left element of a $\GW$-distributed matrix is restricted to a Gamma distribution, which does not contain noncentral chi-squared distributions.
In particular, taking $\v \Sigma = \L \L^T$ to be the cholesky decomposition of $\v \Sigma$, any $\GW$-distributed matrix,
\begin{equation}
    \wM' = (\L \T') (\L \T')^T \sim \GW\b{\v \Sigma, \nu', \boldsymbol \alpha', \boldsymbol \beta', \boldsymbol \mu', \boldsymbol \sigma^{'2}},
\end{equation}
can be written according to the generalised Bartlett decomposition in Eq. \ref{eq:generalised_bartlett}. To obtain the top-left element of $\wM'$, we first explicitly write down the first element of $\L \T'$,
\begin{align}
  \L \T' &= \begin{pmatrix} 
    L_{11} & 0 \\
    L_{21} & L_{22}
  \end{pmatrix}
  \begin{pmatrix}
    T'_{11} \\ T'_{21}
  \end{pmatrix}
  = \begin{pmatrix}
    L_{11} T'_{11} \\ \dotsc
  \end{pmatrix},
\end{align}
where we avoid writing down the second element because it will not be needed.
Then, we can explicitly write down the top-left element of the $\GW$ distributed $\wM'$,
\begin{align}
  \nonumber
  \wM' &= \L \T' (\L \T')^T = 
  \begin{pmatrix}
    L_{11} T'_{11} \\ \dotsc
  \end{pmatrix}
  \begin{pmatrix}
    L_{11} T'_{11} \\ \dotsc
  \end{pmatrix}^T\\
  &= 
  \begin{pmatrix}
    L_{11}^2 (T'_{11})^2 & \dotsc\; \\ \dotsc\; & \dotsc\;
  \end{pmatrix}.
\end{align}
By Eq. \ref{eq:generalised_bartlett_diagonal}, we have $(T_{11}')^2 \sim \gammad{\alpha_1',\beta_1'}$, and so the top-left element of $\wM$ has distribution 
\begin{equation}
    W'_{11} = L_{11}^2(T_{11}')^2 \sim \gammad{\alpha_1',\frac{\beta_1'}{L_{11}^2}}.
\end{equation}
Since the Gamma distribution is not capable of capturing noncentral chi-square distributions, we conclude that the $\AGW$ family is strictly larger than the $\GW$ family.

Figure \ref{fig:qq_hist} shows a probability plot to empirically demonstrate that the $\GW$ distribution is not capable of capturing the $\AGW$ distribution. Specifically, we consider the same $\AGW$ top-left element as in our counter-example above, using $\mu=3, \sigma^2=1$ in Eq. \ref{eq:counter_T}, and take samples from it. We then fit a Gamma distribution to these samples, and show that there is a clear mismatch.
\begin{algorithm}[tb]
\caption{Computing predictions/ELBO for a DWP with $\ABGW$ variational posterior.}
\label{alg:alg1}
\begin{algorithmic}
  \STATE {\bfseries Hyperparameters:} $\{\sl{\nu_\ell}\}_{\ell=1}^L$
  \STATE {\bfseries Learned $\operatorname{Q}$ parameters:}
  \STATE\quad$\{\A_\ell', \B_\ell, \v V_\ell, \v{\alpha}_\ell, \v{\beta}_\ell, \v{\mu}_\ell, \v{\sigma}_\ell, q_\ell\}_{\ell=1}^L$
  \STATE {\bfseries Other learned parameters:} $\X_{\text{i}}$
  \STATE {\bfseries Inputs:} $\X_\text{t}$; {\bfseries Targets:} $\Y$
  \STATE
  \STATE $\X = \begin{pmatrix} \X_\text{i} & \X_\text{t} \end{pmatrix}$ 
  \STATE $\G_0 = \tfrac{1}{\nu_0} \X \X^T$
  \FOR{$\ell$ {\bfseries in} $\{1,\dotsc,L\}$}
    \STATE $\S \leftarrow \tfrac{1}{\nu_\ell} \K(\G_{\ell-1})$
    \STATE\textcolor{gray}{apply Eq.~\eqref{eq:funny_A_equation}}
    \STATE ${\v A}_{l} \leftarrow \text{chol}\b{(1-q_\ell) {\v S_{\text{ii}}} + q_\ell \V_\ell {\V}_\ell^T}{\v A_\ell}'$
    \STATE \textcolor{gray}{sample inducing Gram matrix}
    \STATE $\left(\A_\ell \T_\ell \B_\ell\right) \left(\A_\ell \T_\ell \B_\ell\right)^T = \G_\text{ii}^\ell \sim \Q{\G_\text{ii}^\ell| \G_\text{ii}^{\ell-1}}$ 
    \STATE \textcolor{gray}{update ELBO}
    \STATE $\mathcal{L} \leftarrow \mathcal{L} + \log \P{\G_\text{ii}^\ell| \G_\text{ii}^{\ell-1}} - \log \Q{\G_\text{ii}^\ell| \G_\text{ii}^{\ell-1}}$ 
    \STATE \textcolor{gray}{sample full Gram matrix from conditional prior}
    \STATE $\S_{\text{tt}\cdot \text{i}} \leftarrow \S_\text{tt} - \S^T_\text{it} \S_\text{ii}^{-1} \S_\text{it}$
    \STATE $\F_\text{i}^\ell \leftarrow \A_\ell \T_\ell \B_\ell$
    \STATE $\F^\ell_\text{t} \sim \MN{\S_\text{ti}^T \S_\text{ii}^{-1} \F_\text{i}, \S_{\text{tt}\cdot \text{i}}, \I}$
    \STATE $\G_\ell = 
    \begin{pmatrix}
      \G_\text{ii}^\ell & \F^\ell_\text{i} (\F^\ell_\text{t})^T\\
      \F^\ell_\text{t} (\F^\ell_\text{i})^T & \F^\ell_\text{t} (\F^\ell_\text{t})^T
    \end{pmatrix}$
  \ENDFOR
  \STATE \textcolor{gray}{sample GP inducing outputs and update ELBO}
  \STATE $\F_\text{i}^{L+1} \sim \Q{\F_\text{i}^{L+1} | \G^L_\text{ii}}$
  \STATE $\mathcal{L} \leftarrow \mathcal{L} + \log \P{\F_\text{i}^{L+1}| \G_\text{ii}^L} - \log \Q{\F_\text{i}^{L+1}| \G_\text{ii}^L}$ 
  \STATE \textcolor{gray}{sample GP predictions conditioned on inducing points}
  \STATE $\F_\text{t}^{L+1} \sim \Q{\F_\text{t}^{L+1} | \G^L, \F_\text{i}^{L+1}}$
  \STATE \textcolor{gray}{add likelihood to ELBO}
  \STATE $\mathcal{L} \leftarrow \mathcal{L} + \log \P{\Y| \F_\text{t}^{L+1}}$
\end{algorithmic}
\end{algorithm}

\section{Results}

To compare our $\AGW$ and $\ABGW$ approximate posteriors to the $\GW$ approximate posterior from~\cite{ober2021vardwp}, we trained multiple DWPs on UCI datasets~\citep{pmlr-v48-gal16} using the same architectures, only varying the approximate posterior.
The algorithm for a DWP with an $\ABGW$ approximate posterior is shown in Algorithm~\ref{alg:alg1}. The algorithm for a DWP with an $\AGW$ approximate posterior is recovered by fixing $\v B_\ell = \v I$. The algorithm is similar to Algorithm 1 from~\cite{ober2021vardwp}, but the step for sampling the inducing Gram matrix has changed (since we are using a different approximate posterior).

We also trained DGPs with the same architectures, where we used global inducing point methods from ~\cite{ober21globalinducing}. 
All models were trained with $20\,000$ gradient steps using the ADAM optimizer~\citep{kingma14adam}, with no pre-processing of the data other than normalizing inputs and outputs. An initial learning rate of $10^{-2}$ was used, and after $10\,000$ steps it was set to $10^{-3}$.
RMSE, ELBO and log likelihood are all reported plus or minus one standard error, calculated over 20 splits (apart from the \protein~dataset, where 5 splits were used). 
Results are shown for a 5-layer architecture in Table~\ref{tab:dwp:uci_comb}, and results for 2, 3, and 4 layers can be found in \appendixDetailedResults. Layer widths $\nu_l$ in all layers were set to the number of features in the input data.

Taking the standard errors into account, we see that $\AGW$ and $\ABGW$ approximate posteriors are uniformly as good or better than $\GW$ approximate posteriors across all metrics in the 5-layer case, and this is the case for almost all the experiments we ran (see \appendixDetailedResults). 
Notably, $\AGW$ and $\ABGW$ approximate posteriors are able to achieve higher ELBO and log likelihoods (this is expected since they provide more flexible approximate posteriors).
The largest improvements in ELBO are for \yacht~and \naval,
and in the case of \yacht~this leads to a large gain in RMSE.

Note that the dataset size varies, with the smallest being \yacht~with 308 observations, and the largest being \protein~with 45730 observations.
Training times per epoch for a large and a small dataset, \protein~and \boston~(506 observations) can be found in Table~\ref{tab:dwp:runtime}.
The results show that training time for the models with $\AGW$ and $\ABGW$ approximate posteriors is very similar to that of previous DWPs with $\GW$ approximate posteriors in~\citet{ober2021vardwp}, so the additional computational cost incurred by adding the new parameters $\A$ and $\B$ is negligible. All DWP models tested trained faster than the equivalent DGP models.

\section{Conclusion}
We extended the generalised (singular) Wishart distribution, $\GW$, introduced by \cite{ober2021vardwp} to the $\AGW$
and $\ABGW$ distributions, which we proved (both analytically and empirically) to be strictly more flexible than the $\GW$ distribution. These A- and AB-generalisations of the Wishart distribution are effective when used as approximate posteriors for DWPs, as shown by the near-universal improvement in predictive performance on UCI datasets, both over similar DGP models, and over DWP models that use the less flexible $\GW$ distribution for their approximate posteriors. Furthermore, we showed that this increased flexibility comes at a negligible additional cost in computation. As a result, this is the first DWP work to achieve equal-or-better predictive performance than comparable DGPs on UCI datasets (and it is also cheaper to train). This is significant as DGP priors are equivalent to DWP priors \citep{aitchison2021dkp}, but DWP posteriors are invariant to certain types of posterior symmetries that affect DGPs, meaning they should in theory be easier to capture under variational inference, but until this work practical results had not shown this to be the case.

\bibliography{uai2023-template}

\ifthenelse{\equal{\includeappendix}{yes}}{
  \onecolumn
\appendix
\title{An Improved Variational Approximate Posterior for the Deep Wishart Process\\(Supplementary Material)}
\maketitle
\section{Derivation of $\AGW$ and $\ABGW$ densities}\label{app:densities_deriv}
We briefly provide further background on the Wishart distribution, the Barlett decomposition, and discuss how to derive Jacobians for matrix transformations. Then we use this machinery to derive densities for the $\AGW$ and $\ABGW$ distributions.
\subsection{The Wishart distribution}
The Wishart distribution, $\Wish{\S, \nu}$, is a distribution over positive semi-definite $P \times P$ matrices, where  $\S\in\R^{P\times P}$ is a positive definite covariance matrix, and $\nu > 0$ is an integer-valued degrees-of-freedom parameter. The Wishart distribution is most straightforwardly interpreted as a sum of outer products of multivariate Gaussian random variables. That is, if
we define a random matrix $\wM$ such that,
\begin{align}
  \v f_\lambda &\siid \N{\0, \S},\,\lambda\in\{1,\ldots,\nu\},\\
  \wM &= \sum_{\lambda=1}^\nu \v f_\lambda \v{f}^T_\lambda,
\end{align}
then we say that $\wM$ is Wishart distributed, and write $\wM \sim \Wish{\S, \nu}$. Equivalently, $\wM = \F \F^T $, where $\F\in\R^{P\times P}$ is defined by stacking the vectors $\v f_\lambda$, $\F = \b{\v f_1\quad \cdots\quad\v f_\lambda}$. We say that $\wM$ is standard Wishart distributed if $\S = \I$.

It is easy to generate Wishart random matrices from only standard Gaussian samples. Take $\L = \chol(\S)$ to be the Cholesky of $\S$ and $\v \xi_\lambda \siid\N{\0, \I}$, then  $\L\v\xi_\lambda \siid \N{\0, \S}$. It follows that,
\begin{align}\label{eq:wishart_as_std_normal}
\L\b{\sum_{\lambda=1}^\nu \v \xi_\lambda \v \xi_\lambda^T} \L^T = \L \v \Xi \v \Xi^T \L^T \sim \Wish{\S, \nu},
\end{align}
where $\v \Xi$ is the matrix of stacked vectors $\v \xi_\lambda$ such that $\v \Xi = \b{\v \xi_1\quad\cdots\quad \v \xi_\nu}$. From~\eqref{eq:wishart_as_std_normal}, it can be observed that
${\E\sqb{\wM} = \L (\nu \I)\L^T = \nu\S}$. Additionally, $\H=\v \Xi \v \Xi^T$ is standard Wishart distributed, therefore~\eqref{eq:wishart_as_std_normal} also
gives us a way to transform a standard Wishart into a Wishart with covariance parameter $\S$: $\H \sim \Wish{\I, \nu}\implies \L \H \L^T\sim\Wish{\S, \nu}$.

Finally, note that the density of the Wishart distribution is
given by,
\begin{align}
    \label{eq:dwp:wishdens}
    \pt\b{\wM} = \frac{\pi^{\nu (\nt - P)/2}}{2^{\nu P/2}\abs{\S}^{\nu/2} \v \Gamma_\nt\b{\tfrac{\nu}{2}}} \abs{\wM_{:\nt, :\nt}}^{\b{\nu-P-1}/2} \etr \b{-\S^{-1}\wM/2},
\end{align}
where $\nt = \min(\nu, P)$, and $\Gamma_\nt$ is the multivariate gamma function~\citep{srivastava2003singularwishart}.
\subsection{The Bartlett decomposition and some generalisations}
Suppose $\wM\sim \Wish{\I,\nu}$, and $\nu\geq P$, then 
the Bartlett decomposition \citep{bartlett1933on} allows for efficient sampling of $\wM$ (the constraint $\nu\geq P$ refers to the fact that $\wM$ almost surely has full rank). Rather than sampling $\nu P^2$ Gaussian random variables to construct $\wM$ (which can become prohibitively costly when $\nu$ is large),
the Bartlett decomposition allows us to sample only $P(P-1)/2$ Gaussian random variables, and $P$ Gamma random variables. In particular, if $\T$ is a random matrix distributed according to,
\begin{subequations}\label{eq:standard_bartlett}
\begin{align}
\T &= \begin{pmatrix}
  T_{11} & \dotsm & 0 \\
  \vdots & \ddots & \vdots \\
  T_{P1} & \dotsm & T_{PP}
\end{pmatrix},\\
\pt\b{T_{jj}^2} &= \gammat\b{T_{jj}^2;\, \tfrac{\nu-j+1}{2}, \tfrac{1}{2}},\\
\pt\b{T_{j > k}} &= \Nc{T_{jk}}{\,0, 1},
\end{align}
\end{subequations}
then $\T\T^T \sim \Wish{\I,\nu}$.
The utility of~\eqref{eq:standard_bartlett} can be extended in two ways. Firstly, we can use~\eqref{eq:standard_bartlett} to sample from non-standard Wisharts, since $\L (\T \T^T) \L^T \sim\Wish{\L \L^T, \nu}$. Secondly,~\cite{srivastava2003singularwishart} extends the Bartlett decomposition to allow for sampling of singular Wisharts. Suppose $\nu < P$, and take $\T$ to be distributed according to,
\begin{subequations}\label{eq:nonsingular_bartlett}
\begin{align}
  \T &= \begin{pmatrix}
    T_{11}    & \dotsm & 0          \\ 
    \vdots    & \ddots & \vdots     \\ 
    T_{\nu 1} & \dotsm & T_{\nu \nu} \\ 
    \vdots    & \ddots & \vdots     \\ 
    T_{\nD 1}   & \dotsm & T_{\nD \nu}
  \end{pmatrix},\\
  \pt\b{T_{ii}^2} &= \gammad{T_{ii}^2;\, \tfrac{\nu-j+1}{2}, \tfrac{1}{2}},\,  i \in\{1,\ldots,\nu\},\\
  \pt\b{T_{i > j}} &=\normal{T_{i>j};\,0, 1},
\end{align}
\end{subequations}
then $\T \T^T \sim \Wish{\I, \nu}$.

We arrive at the A- and AB-generalised (singular) Wishart distributions by generalising the (singular) Barlett decomposition in~\eqref{eq:nonsingular_bartlett}.
Concretely, we borrow the form of~\eqref{eq:nonsingular_bartlett}, but allow the parameters of the Gaussian and gamma distributions to be arbitrary,
\begin{subequations}\label{eq:generalised_bartlett2}
\begin{align}
  \T &= \begin{pmatrix}
    T_{11}    & \dotsm & 0          \\ 
    \vdots    & \ddots & \vdots     \\ 
    T_{\nu 1} & \dotsm & T_{\nu \nu} \\ 
    \vdots    & \ddots & \vdots     \\ 
    T_{\nD 1}   & \dotsm & T_{\nD \nu}
  \end{pmatrix},\\
  \pt\b{T_{ii}^2} &= \gammad{T_{ii}^2;\, \alpha_i,\beta_i},\,  i \in\{1,\ldots,\nu\},\\
  \pt\b{T_{i > j}} &=\normal{T_{i>j};\,\mu_{ij}, \sigma^2_{ij}}.
\end{align}
\end{subequations}
For any invertible matrix $\A\in\R^{P\times P}$ and any invertible lower triangular $\B \in \R^{\nu \times \nu}$, we write ${\A \T \T^T \A^T\sim \AGW\b{\A, \nu, \v \alpha, \v \beta, \v \mu, \v \sigma}}$ and $\A \T \B \B^T \T^T \A^T\sim \ABGW\b{\A, \B, \nu, \v \alpha, \v \beta, \v \mu, \v \sigma}$.
Given the necessary parameters, it is straightforward to sample
matrices from the $\AGW$ and $\ABGW$ families using~\eqref{eq:generalised_bartlett2}. However it is non-trivial to write down the corresponding densities --- the rest of this section is dedicated to this task.
\subsection{Jacobians for matrix transformations}\label{sec:app:jac:deriving}
We want to obtain the densities of $\wM_A := \A \T \T^T \A^T$ and $\wM_{AB} := \A \T \B \B^T \T^T\A^T$, where we know the density of $\T$.
Ultimately, we will use the change of variables formula,
\begin{align}
    \qd{\wM} = \qd{\T}\abs{\dd[\T]{\wM}},
\end{align}
where $\abs{\partial{\T} / \partial{\wM}}$ is the Jacobian determinant of the transformation.

For a vector-vector transformation $\v y = \v f(\v x)$, where $\v x\in\R^n$ and $\v y \in \R^{m}$, the Jacobian $\partial \v y/\partial \v x$ can be calculated by
evaluating $\partial y_i/\partial x_j$ for $i\in\{1,\ldots,m\}$, and $j\in\{1,\ldots,n\}$. It is less simple to calculate the Jacobian for matrix-matrix transformations, but it can be done by stacking the columns of our matrices into a long vector, and then calculating the associated vector-vector Jacobian. We demonstrate this
with a simple example for $2\times 2$ matrices. Consider,
\begin{align}
    \underbrace{\begin{pmatrix} Y_{11} & Y_{12} \\ Y_{21} & Y_{22} \end{pmatrix}}_{\Y} &=
    \underbrace{\begin{pmatrix} A_{11} & A_{12} \\ A_{21} & A_{22} \end{pmatrix}}_{\mathbf{A}}
    \underbrace{\begin{pmatrix} X_{11} & X_{12} \\ X_{21} & X_{22} \end{pmatrix}}_{\X}.
\end{align}
We `vectorise' $\Y$ and $\X$ to obtain,
\begin{align}
    \begin{pmatrix}
      Y_{11} \\
      Y_{21} \\
      Y_{12} \\
      Y_{22}
    \end{pmatrix}
    &=
    \underbrace{\begin{pmatrix}
      A_{11} & A_{12} & 0 & 0 \\
      A_{21} & A_{22} & 0 & 0 \\
      0 & 0 & A_{11} & A_{12} \\
      0 & 0 & A_{21} & A_{22}
    \end{pmatrix}}_{\mathbf{A}^*}
    \begin{pmatrix}
      X_{11} \\
      X_{21} \\
      X_{12} \\
      X_{22}
    \end{pmatrix}.
\end{align}
The Jacobian of this transformation is clearly,
\begin{align}
\dd[\Y]{\X} = \A^*,
\end{align}
and the associated Jacobian determinant is therefore,
\begin{align}
\abs{\dd[\Y]{\X}} = \abs{\A}^2.
\end{align}
We now consider how to calculate some Jacobian determinants that are relevant in calculating the $\AGW$ and $\ABGW$ densities.
\subsection{Jacobian for the product of a lower triangular matrix with itself}\label{sec:jac:J_LLT}
Consider the transformation $\G = \v \La \La^T$, where $\La \in \reals^{P \times P}$, and $\La$ is lower triangular.~\cite{ober2021vardwp} showed that the Jacobian determinant is,
\begin{align}
  \label{eq:jac:J_LaLaT}
  \abs{\dd[\G]{\La}} = \prod_{i=1}^P 2 \Lambda_{ii}^{P - i + 1}.
\end{align}
They also showed that the same transformation, $\G = \v \La \La^T$, but in the case $\La\in\R^{P\times \nu}$
has Jacobian determinant,
\begin{align}\label{eq:jac:J_LaLaTsing}
  \abs{\dd[\G]{\La}} &= \prod_{i=1}^\nt 2 \Lambda_{ii}^{P - i + 1},
\end{align}
where $\tilde\nu = \min\{P, \nu\}$.
\subsection{Jacobian for the product of two different lower triangular matrices}\label{sec:jac:J_LA}
Consider the transformation $\T \mapsto \La = \L \T$, where $\T \in \reals^{P \times \nu}$ and is lower triangular, and $\L\in\R^{P\times P}$ is also lower triangular.
~\cite{ober2021vardwp} showed that the Jacobian determinant is,
\begin{align}
\label{eq:jac:J_LA}
  \abs{\dd[\v \La]{\T}} = \prod_{i=1}^P L_{ii}^{\min(i, \nu)}.
\end{align}
We also need the Jacobian determinant for a right linear transformation. Therefore, consider also the transformation $\T\mapsto\La = \T\B$, where again $\T\in\reals^{P\times \nu}$,
but $\B\in\reals^{\nu\times \nu}$ and is invertible lower triangular. It is helpful to write down the matrices explicitly,
\begin{align}
  \label{eq:jac:AL}
  \begin{pmatrix}
    \Lambda_{11}    & \dotsm & 0          \\ 
    \vdots          & \ddots & \vdots     \\ 
    \Lambda_{\nu 1} & \dotsm & \Lambda_{\nu \nu} \\ 
    \vdots          & \vdots & \vdots     \\ 
    \Lambda_{P 1}   & \dotsm & \Lambda_{P \nu}
  \end{pmatrix} = 
    \begin{pmatrix}
    T_{11}    & \dotsm & 0          \\ 
    \vdots          & \ddots & \vdots     \\ 
    T_{\nu 1} & \dotsm & T_{\nu \nu} \\ 
    \vdots          & \vdots & \vdots     \\ 
    T_{P 1}   & \dotsm & T_{P \nu}
  \end{pmatrix}
  \begin{pmatrix}
  B_{11} & \dotsm & 0 \\
  \vdots & \ddots & \vdots \\
  B_{\nu 1} & \dotsm & B_{\nu\nu}
  \end{pmatrix},
\end{align}
and consider the rows of $\La$. For the first row, we have
\begin{align*}
    \begin{pmatrix}
      \Lambda_{11} 
    \end{pmatrix}
    = 
    \begin{pmatrix}
      T_{11}
    \end{pmatrix}
    \begin{pmatrix}
      B_{11}
    \end{pmatrix},
\end{align*}
or equivalently,
\begin{align*}
    \La_{1, :1} = \T_{1, :1}\B_{:1, :1}.
\end{align*}
Similarly, for rows up to the $\nu^\text{th}$ row, i.e.\ for $i \leq \nu$, we have,
\begin{align*}
    \begin{pmatrix}
      \Lambda_{i1} & \dotsm & \Lambda_{ii}
    \end{pmatrix}
    =
    \begin{pmatrix}
      T_{i1} & \dotsm & T_{ii}
    \end{pmatrix}
    \begin{pmatrix}
      B_{11} & \dotsm & 0 \\
      \vdots & \ddots & \vdots \\
      B_{i1} & \dotsm & B_{ii}
    \end{pmatrix},
\end{align*}
which can be written as
\begin{align*}
    \La_{i, :i} = \T_{i, :i}\B_{:i, :i}.
\end{align*}
For rows beyond the $\nu^\text{th}$ row, i.e., $i > \nu$, the expression becomes,
\begin{align*}
    \begin{pmatrix}
      \Lambda_{i1} & \dotsm & \Lambda_{i\nu}
    \end{pmatrix}
    =
    \begin{pmatrix}
      T_{i1} & \dotsm & T_{i\nu}
    \end{pmatrix}
    \begin{pmatrix}
      B_{11} & \dotsm & 0 \\
      \vdots & \ddots & \vdots \\
      B_{\nu1} & \dotsm & B_{\nu \nu}
    \end{pmatrix},
\end{align*}
which again can be written as,
\begin{align*}
    \La_{i, :\nu} = \T_{i, :\nu} \B_{:\nu, :\nu} = \T_{i, :} \B.
\end{align*}
To calculate the Jacobian, we proceed by taking the transpose of each of the rows and stacking them, giving,
\begin{align*}
    \begin{pmatrix}
      \La_{1, :1}^\top \\
      \La_{2, :2}^\top \\
      \vdots \\
      \La_{\nu, :\nu}^\top \\
      \La_{\nu + 1, :\nu}^\top \\
      \vdots \\
      \La_{P, :\nu}^\top
    \end{pmatrix}
    =
    \begin{pmatrix}
      \B_{:1, :1}^\top & \0 & \dotsm & \0 & \0 & \dotsm & \0 \\
      \0 & \B_{:2, :2}^\top & \dotsm & \0 & \0 & \dotsm & \0 \\
      \vdots & \vdots & \ddots & \vdots & \vdots & \ddots & \vdots \\
      \0 & \0 & \dotsm & \B^\top & \0 & \dotsm & \0 \\
      \0 & \0 & \dotsm & \0 & \B^\top & \dotsm & \0 \\
      \vdots & \vdots & \ddots & \vdots & \vdots & \ddots & \vdots \\
      \0 & \0 & \dotsm & \0 & \0 & \dotsm & \B^\top \\
    \end{pmatrix}
    \begin{pmatrix}
      \T_{1, :1}^\top \\
      \T_{2, :2}^\top \\
      \vdots \\
      \T_{\nu, :\nu}^\top \\
      \T_{\nu + 1, :\nu}^\top \\
      \vdots \\
      \T_{P, :\nu}^\top
    \end{pmatrix}.
\end{align*}
Since the square matrix is upper triangular, its determinant is simply the product of the elements of its diagonal. This gives the Jacobian determinant,
\begin{align}\label{eq:jac:TB}
\abs{\dd[\La]{\T}} =  \prod_{i=1}^{\nt} B_{ii}^{P - i + 1}.
\end{align}
\subsection{Jacobian for $\C = \La\La^\top \mapsto \D = \A\C\transpose{\A}$, where $\A$ is an invertible matrix}
Now consider the transformation $\C = \La\La^\top\mapsto \D = \A\C\A^T$, where $\La \in \reals^{P\times \nu}$ is lower triangular with rank $\nu$, and $\A\in\R^{P\times P}$ is invertible. This Jacobian is difficult to derive from scratch; however, we can obtain it using the density of the singular Wishart.
In particular, the probability density function of $\D \sim \mathcal{W}(\S, \nu)$ is given by,
\begin{align*}
    \pt_{1}(\D) = \frac{\pi^{\nu(\nt - P)/2}}{2^{\nu P/2}\abs{\S}^{\nu/2}\Gamma_{\nt}\b{\tfrac{\nu}{2}}} \abs{\D_{:\nt, :\nt}}^{(\nu - P - 1)/2}\etr \b{-\S^{-1}\D /2},
\end{align*}
where $\nt = \min \b{\nu, P}$ as before. 
Note that $\D_{:\nt, :\nt}$ is almost surely full rank.
For $\C \sim \mathcal{W}(\I_P, \nu)$, this simplifies to,
\begin{align*}
    \pt_{2}(\C) = \frac{\pi^{\nu(\nt-P)/2}}{2^{\nu P/2}\Gamma_{\nt}\b{\tfrac{\nu}{2}}}\abs{\C_{:\nt,:\nt}}^{(\nu-P-1)/2}\etr\b {-\C/2}.
\end{align*}
Using these densities, we can use the identity,
\begin{align*}
    \pt_1(\D) = \pt_2\b{\C} \abs{\dd[\C]{\D}},
\end{align*}
to obtain the desired Jacobian determinant,
\begin{align}\label{eq:jac:J_AXAT}
    \abs{\dd[\D]{\C}} = \pt_2\b{\C}/\pt_1(\D) = 
    \abs{\A \A^T}^{\nu/2} \frac{\abs{\C_{:\nt,:\nt}}^{(\nu-P-1)/2}}{\abs{\D_{:\nt,:\nt}}^{(\nu - P - 1)/2}} = \abs{\A}^{\nu} \frac{\abs{\C_{:\nt,:\nt}}^{(\nu-P-1)/2}}{\abs{\D_{:\nt,:\nt}}^{(\nu - P- 1)/2}}.
\end{align}
We can now put these Jacobian determinant results together to derive the densities for the A- and AB-generalised (singular) Wishart distributions.
\subsection{The A-generalised (singular) Wishart density}\label{appendix:deriving_densities}
In Section~\ref{sec:a_ab_gswd} we said that $\G = \A \T (\A \v T)^T\sim\AGW\b{\A, \nu, \v \alpha, \v \beta, \v \mu, \v \sigma}$ if
$\A\in\R^{P\times P}$ is invertible and $\T\in\R^{P\times \nu}$ is distributed according to~\eqref{eq:generalised_bartlett}. If we define $\C$ such that $\G = \A \C \A^T = \A \T \T^T \A^T$, then by the change of variables formula for probability densities,
\begin{align}
    \qd{\G} = \qd{\T}\abs{\dd[\T]{\C}}\abs{\dd[\C]{\G}}.
\end{align}
By combining the density of $\T$,
\begin{align}
\qd{\T} = 2^{\nt}
\prod_{j=1}^\nt T_{jj}\gammat\b{T_{jj}^2;\,\alpha_j, \beta_j}\prod_{i=j+1}^P \Nc{T_{ij}}{\mu_{ij}, \sigma_{ij}^2},
\end{align}
the result from~\eqref{eq:jac:J_LaLaTsing},
\begin{align}
  \abs{\dd[\C]{\T}} &= 2^\nt\prod_{j=1}^\nt  T_{jj}^{P - j + 1},
\end{align}
and the result from~\eqref{eq:jac:J_AXAT},
\begin{align}
    \abs{\dd[\G]{\C}} = 
     \abs{\A}^{\nu} \frac{\abs{\C_{:\nt,:\nt}}^{(\nu-P-1)/2}}{\abs{\G_{:\nt,:\nt}}^{(\nu - P - 1)/2}},
\end{align}
we obtain the A-generalised (singular) Wishart density,
\begin{align}
    \qd{\G} &= \frac{\abs{\G_{:\nt, :\nt}}^{(\nu - P - 1)/2}}{\abs{\A}^\nu\abs{\C_{:\nt, :\nt}}^{(\nu - P- 1)/2}} 
    \prod_{j=1}^\nt \frac{\gammat\b{T_{jj}^2;\,\alpha_j, \beta_j}}{T_{jj}^{P-j}}\prod_{i=j+1}^P \Nc{T_{ij}}{\mu_{ij}, \sigma_{ij}^2}.
\end{align}
\subsection{The AB-generalised (singular) Wishart density}
The derivation for the AB-generalised (singular) Wishart is similar to that of the A-generalised (singular) Wishart, with the addition of one extra step.
Namely, as the AB-generalised (singular) Wishart defines $\G = \A\T\B\transpose{\b{\A\T\B}}$, we define $\La = \T\B$ and $\C = \La\transpose{\La}$, so that,
\begin{align*}
    \qd{\G} = \qd{\T}\abs{\dd[\T]{\La}}\abs{\dd[\La]{\C}}\abs{\dd[\C]{\D}}.
\end{align*}
This first Jacobian determinant can be obtained using~\eqref{eq:jac:TB}, 
\begin{align*}
    \abs{\dd[\T]{\La}} = \prod_{i=1}^\nt \frac{1}{B_{ii}^{P - i + 1}},
\end{align*}
whereas the second,
\begin{align*}
    \abs{\dd[\La]{\C}} &= \frac{1}{2^\nt}\prod_{i=1}^{\nt} \frac{1}{\Lambda_{ii}^{P - i + 1}}= \frac{1}{2^\nt}\prod_{i=1}^{\nt} \frac{1}{T_{ii}^{P - i + 1}B_{ii}^{P - i + 1}},
\end{align*}
arises from~\eqref{eq:jac:J_LaLaTsing}.
The remaining Jacobians remain unchanged in form, so that our final density is given by,
\begin{align}
    \qd{\G} &= \frac{\abs{\G_{:\nt, :\nt}}^{(\nu - P - 1)/2}}{\abs{\A}^\nu\abs{\C_{:\nt, :\nt}}^{(\nu - P - 1)/2}} 
    \prod_{j=1}^\nt \frac{\gammat\b{T_{jj}^2;\,\alpha_j, \beta_j}}{T_{jj}^{P-j}B_{jj}^{2(P-j+1)}}\prod_{i=j+1}^P \Nc{T_{ij}}{\mu_{ij}, \sigma_{ij}^2}.
\end{align}
\section{Detailed Experimental Results}\label{appendix:detailed_results}
All models were trained on the UCI splits from~\cite{pmlr-v48-gal16}, of which there are 20 for each dataset apart from \protein. The datasets and the splits are available at~\url{https://github.com/yaringal/DropoutUncertaintyExps/tree/master/UCI_Datasets}. Deep Wishart processes
with the three kinds of approximate posterior ($\GW$, $\AGW$, and $\ABGW$)  were trained, with number of layers $\ell\in\{2,\ldots,5\}$, and width $\nu_\ell$ fixed to the number of input features. We applied the squared exponential kernel as a non-linearity at each layer,
with automatic relevance determination (ARD,~\cite{williams2006gaussian}) in the first layer only. The DGPs trained reflected this architecture, with each GP layer returning features with dimension equal to the number of input features. In particular the DGPs were trained using global inducing point methods~\citep{ober21globalinducing}. The final layer of the DWP also uses
a global inducing approximate posterior~\citep{ober21globalinducing}.

All models were trained using the same scheme. $20\,000$ gradient steps were used to train each model, with the ADAM optimizer~\cite{kingma14adam}.
We began with an initial learning rate of $10^{-2}$, and then stepped the learning rate down to $10^{-3}$ after $10\,000$ gradient steps. The KL was annealed using a factor increasing linearly from $0$ to $1$ over the first $1\,000$ gradient steps. No pre-processing of the data was performed, other than normalizing inputs and outputs. To train, $10$ samples were drawn from the approximate posterior, and to test $100$ samples were drawn. For the smaller datasets (\boston, \concrete, \energy, \wine, \yacht), training was performed on a CPU (Intel Core i9-10900X), and for the other (larger) datasets, an internal cluster of machines was used, with NVIDIA GeForce 2080 Ti GPUs.
\subsection{Tables}
\cref{tab:dwp:uci_elbos1,tab:dwp:uci_elbos2,tab:dwp:uci_lls,tab:dwp:uci_rmses} report the ELBOs, test log likelihoods, and RMSEs from our UCI experiments respectively. In all cases, we give the mean of each metric (plus or minus one standard error), and highlight the model with the best mean value in bold for each configuration (unless all are equal).
\begin{table}[ht]
\footnotesize
  \caption{ELBOs per datapoint. We report mean plus or minus one standard error over the splits. Bold numbers correspond to the best models overall.}
  \label{tab:dwp:uci_elbos1}
  \centering
  \begin{tabular}{rcccc}
    \toprule
& & & DWP & \\
\{Dataset\}-\{Depth\} & DGP & $\QGW$ & $\QAGW$ & $\QABGW$ \\
\midrule  
\textsc{\boston} - 2 & -0.38 $\pm$ 0.01 & -0.33 $\pm$ 0.00 & \textbf{-0.32 $\pm$ 0.01} & \textbf{-0.32 $\pm$ 0.00} \\ 
3 & -0.40 $\pm$ 0.00 & -0.34 $\pm$ 0.01 & \textbf{-0.33 $\pm$ 0.00} & \textbf{-0.33 $\pm$ 0.01} \\ 
4 & -0.43 $\pm$ 0.00 & \textbf{-0.35 $\pm$ 0.00} & \textbf{-0.34 $\pm$ 0.01} & \textbf{-0.34 $\pm$ 0.01} \\ 
5 & -0.45 $\pm$ 0.00 & \textbf{-0.37 $\pm$ 0.01} & \textbf{-0.36 $\pm$ 0.00} & \textbf{-0.36 $\pm$ 0.00} \\ 
\midrule 
\textsc{\concrete} - 2 & -0.45 $\pm$ 0.00 & -0.42 $\pm$ 0.00 & -0.40 $\pm$ 0.00 & \textbf{-0.39 $\pm$ 0.00} \\ 
3 & -0.47 $\pm$ 0.00 & -0.43 $\pm$ 0.00 & \textbf{-0.41 $\pm$ 0.00} & \textbf{-0.41 $\pm$ 0.00} \\ 
4 & -0.49 $\pm$ 0.00 & -0.46 $\pm$ 0.00 & \textbf{-0.43 $\pm$ 0.00} & \textbf{-0.43 $\pm$ 0.00} \\ 
5 & -0.50 $\pm$ 0.00 & -0.49 $\pm$ 0.00 & \textbf{-0.45 $\pm$ 0.00} & \textbf{-0.45 $\pm$ 0.00} \\ 
\midrule 
\textsc{\energy} - 2 & 1.43 $\pm$ 0.00 & \textbf{1.46 $\pm$ 0.00} & \textbf{1.46 $\pm$ 0.00} & \textbf{1.46 $\pm$ 0.00} \\ 
3 & 1.42 $\pm$ 0.00 & 1.44 $\pm$ 0.00 & \textbf{1.45 $\pm$ 0.00} & \textbf{1.45 $\pm$ 0.00} \\ 
4 & 1.40 $\pm$ 0.00 & 1.42 $\pm$ 0.00 & \textbf{1.43 $\pm$ 0.00} & \textbf{1.43 $\pm$ 0.00} \\ 
5 & 1.38 $\pm$ 0.00 & 1.40 $\pm$ 0.00 & \textbf{1.42 $\pm$ 0.00} & 1.41 $\pm$ 0.00 \\ 
\midrule 
\textsc{\kinnm} - 2 & -0.15 $\pm$ 0.00 & -0.16 $\pm$ 0.00 & \textbf{-0.14 $\pm$ 0.00} & \textbf{-0.14 $\pm$ 0.00} \\ 
3 & -0.14 $\pm$ 0.00 & -0.15 $\pm$ 0.00 & \textbf{-0.13 $\pm$ 0.00} & \textbf{-0.13 $\pm$ 0.00} \\ 
4 & -0.14 $\pm$ 0.00 & -0.14 $\pm$ 0.00 & \textbf{-0.11 $\pm$ 0.00} & \textbf{-0.11 $\pm$ 0.00} \\ 
5 & -0.14 $\pm$ 0.00 & -0.14 $\pm$ 0.00 & \textbf{-0.11 $\pm$ 0.00} & \textbf{-0.11 $\pm$ 0.00} \\ 
\midrule 
\textsc{\naval} - 2 & 3.93 $\pm$ 0.05 & 3.82 $\pm$ 0.09 & 3.80 $\pm$ 0.13 & 3.84 $\pm$ 0.10 \\ 
3 & 3.83 $\pm$ 0.06 & 3.71 $\pm$ 0.12 & 3.86 $\pm$ 0.06 & \textbf{3.99 $\pm$ 0.04} \\ 
4 & \textbf{3.91 $\pm$ 0.05} & 3.66 $\pm$ 0.13 & \textbf{3.75 $\pm$ 0.11} & \textbf{3.85 $\pm$ 0.09} \\ 
5 & \textbf{3.92 $\pm$ 0.04} & 3.59 $\pm$ 0.12 & \textbf{3.97 $\pm$ 0.02} & 3.63 $\pm$ 0.22 \\ 
\midrule 
\textsc{\power} - 2 & 0.03 $\pm$ 0.00 & 0.03 $\pm$ 0.00 & \textbf{0.04 $\pm$ 0.00} & \textbf{0.04 $\pm$ 0.00} \\ 
3 & 0.03 $\pm$ 0.00 & 0.03 $\pm$ 0.00 & 0.03 $\pm$ 0.00 & 0.03 $\pm$ 0.00 \\ 
4 & 0.03 $\pm$ 0.00 & 0.03 $\pm$ 0.00 & 0.03 $\pm$ 0.00 & 0.03 $\pm$ 0.00 \\ 
5 & \textbf{0.03 $\pm$ 0.00} & 0.02 $\pm$ 0.00 & \textbf{0.03 $\pm$ 0.00} & \textbf{0.03 $\pm$ 0.00} \\ 
\midrule 
\textsc{\protein} - 2 & \textbf{-1.06 $\pm$ 0.00} & -1.07 $\pm$ 0.00 & \textbf{-1.06 $\pm$ 0.00} & \textbf{-1.06 $\pm$ 0.00} \\ 
3 & -1.04 $\pm$ 0.00 & -1.04 $\pm$ 0.00 & \textbf{-1.03 $\pm$ 0.00} & \textbf{-1.03 $\pm$ 0.00} \\ 
4 & -1.02 $\pm$ 0.00 & -1.02 $\pm$ 0.00 & \textbf{-1.00 $\pm$ 0.00} & -1.01 $\pm$ 0.00 \\ 
5 & \textbf{-1.00 $\pm$ 0.00} & -1.01 $\pm$ 0.00 & \textbf{-1.00 $\pm$ 0.00} & \textbf{-1.00 $\pm$ 0.00} \\ 
\midrule 
\textsc{\wine} - 2 & -1.18 $\pm$ 0.00 & -1.18 $\pm$ 0.00 & \textbf{-1.18 $\pm$ 0.00} & \textbf{-1.18 $\pm$ 0.00} \\ 
3 & -1.19 $\pm$ 0.00 & \textbf{-1.18 $\pm$ 0.00} & \textbf{-1.18 $\pm$ 0.00} & \textbf{-1.18 $\pm$ 0.00} \\ 
4 & -1.19 $\pm$ 0.00 & \textbf{-1.18 $\pm$ 0.00} & \textbf{-1.18 $\pm$ 0.00} & \textbf{-1.18 $\pm$ 0.00} \\ 
5 & -1.19 $\pm$ 0.00 & -1.19 $\pm$ 0.00 & -1.19 $\pm$ 0.00 & -1.19 $\pm$ 0.00 \\ 
\midrule 
\textsc{\yacht} - 2 & 1.88 $\pm$ 0.03 & 2.02 $\pm$ 0.01 & \textbf{2.07 $\pm$ 0.01} & \textbf{2.07 $\pm$ 0.01} \\ 
3 & 1.62 $\pm$ 0.01 & 1.86 $\pm$ 0.02 & \textbf{2.02 $\pm$ 0.01} & \textbf{2.03 $\pm$ 0.01} \\ 
4 & 1.47 $\pm$ 0.02 & 1.73 $\pm$ 0.02 & \textbf{1.93 $\pm$ 0.01} & 1.91 $\pm$ 0.01 \\ 
5 & 1.46 $\pm$ 0.02 & 1.59 $\pm$ 0.02 & \textbf{1.79 $\pm$ 0.02} & \textbf{1.79 $\pm$ 0.02} \\ 
\bottomrule
  \end{tabular}
\end{table}

\begin{table}[ht]
\footnotesize
  \caption{ELBO differences per datapoint. We report mean differences plus or minus one standard error over the splits.}
  \label{tab:dwp:uci_elbos2}
  \centering
  \begin{tabular}{rccc}
    \toprule
\{Dataset\}-\{Depth\}  & $\QAGW - \QGW$ & $\QABGW - \QGW$ & $\QAGW - \QABGW$ \\
\midrule  
\textsc{\boston} - 2   &   0.01 $\pm$ 0.01 &  0.01 $\pm$ 0.00 &   0.00 $\pm$ 0.01 \\
3                      &   0.01 $\pm$ 0.01 &  0.01 $\pm$ 0.01 &   0.00 $\pm$ 0.01 \\
4                      &   0.01 $\pm$ 0.01 &  0.01 $\pm$ 0.01 &   0.00 $\pm$ 0.01 \\
5                      &   0.01 $\pm$ 0.01 &  0.01 $\pm$ 0.01 &   0.00 $\pm$ 0.00 \\
\midrule 
\textsc{\concrete} - 2 &   0.02 $\pm$ 0.00 &  0.03 $\pm$ 0.00 &  -0.01 $\pm$ 0.00 \\
3                      &   0.02 $\pm$ 0.00 &  0.02 $\pm$ 0.00 &   0.00 $\pm$ 0.00 \\
4                      &   0.03 $\pm$ 0.00 &  0.03 $\pm$ 0.00 &   0.00 $\pm$ 0.00 \\
5                      &   0.04 $\pm$ 0.00 &  0.04 $\pm$ 0.00 &   0.00 $\pm$ 0.00 \\
\midrule 
\textsc{\energy} - 2   &   0.00 $\pm$ 0.00 &  0.00 $\pm$ 0.00 &   0.00 $\pm$ 0.00 \\
3                      &   0.01 $\pm$ 0.00 &  0.01 $\pm$ 0.00 &   0.00 $\pm$ 0.00 \\
4                      &   0.01 $\pm$ 0.00 &  0.01 $\pm$ 0.00 &   0.00 $\pm$ 0.00 \\
5                      &   0.02 $\pm$ 0.00 &  0.01 $\pm$ 0.00 &   0.01 $\pm$ 0.00 \\
\midrule 
\textsc{\kinnm} - 2    &   0.02 $\pm$ 0.00 &  0.02 $\pm$ 0.00 &   0.00 $\pm$ 0.00 \\
3                      &   0.02 $\pm$ 0.00 &  0.02 $\pm$ 0.00 &   0.00 $\pm$ 0.00 \\
4                      &   0.03 $\pm$ 0.00 &  0.03 $\pm$ 0.00 &   0.00 $\pm$ 0.00 \\
5                      &   0.03 $\pm$ 0.00 &  0.03 $\pm$ 0.00 &   0.00 $\pm$ 0.00 \\
\midrule 
\textsc{\naval} - 2    &  -0.02 $\pm$ 0.16 &  0.02 $\pm$ 0.13 &  -0.04 $\pm$ 0.16 \\
3                      &   0.15 $\pm$ 0.13 &  0.28 $\pm$ 0.13 &  -0.13 $\pm$ 0.07 \\
4                      &   0.09 $\pm$ 0.17 &  0.19 $\pm$ 0.16 &  -0.10 $\pm$ 0.14 \\
5                      &   0.38 $\pm$ 0.12 &  0.04 $\pm$ 0.25 &   0.34 $\pm$ 0.22 \\
\midrule 
\textsc{\power} - 2    &   0.01 $\pm$ 0.00 &  0.01 $\pm$ 0.00 &   0.00 $\pm$ 0.00 \\
3                      &   0.00 $\pm$ 0.00 &  0.00 $\pm$ 0.00 &   0.00 $\pm$ 0.00 \\
4                      &   0.00 $\pm$ 0.00 &  0.00 $\pm$ 0.00 &   0.00 $\pm$ 0.00 \\
5                      &   0.01 $\pm$ 0.00 &  0.01 $\pm$ 0.00 &   0.00 $\pm$ 0.00 \\
\midrule 
\textsc{\protein} - 2  &   0.01 $\pm$ 0.00 &  0.01 $\pm$ 0.00 &   0.00 $\pm$ 0.00 \\
3                      &   0.01 $\pm$ 0.00 &  0.01 $\pm$ 0.00 &   0.00 $\pm$ 0.00 \\
4                      &   0.02 $\pm$ 0.00 &  0.01 $\pm$ 0.00 &   0.01 $\pm$ 0.00 \\
5                      &   0.01 $\pm$ 0.00 &  0.01 $\pm$ 0.00 &   0.00 $\pm$ 0.00 \\
\midrule 
\textsc{\wine} - 2     &   0.00 $\pm$ 0.00 &  0.00 $\pm$ 0.00 &   0.00 $\pm$ 0.00 \\
3                      &   0.00 $\pm$ 0.00 &  0.00 $\pm$ 0.00 &   0.00 $\pm$ 0.00 \\
4                      &   0.00 $\pm$ 0.00 &  0.00 $\pm$ 0.00 &   0.00 $\pm$ 0.00 \\
5                      &   0.00 $\pm$ 0.00 &  0.00 $\pm$ 0.00 &   0.00 $\pm$ 0.00 \\
\midrule 
\textsc{\yacht} - 2    &   0.05 $\pm$ 0.01 &  0.05 $\pm$ 0.01 &   0.00 $\pm$ 0.01 \\
3                      &   0.16 $\pm$ 0.02 &  0.17 $\pm$ 0.02 &  -0.01 $\pm$ 0.01 \\
4                      &   0.20 $\pm$ 0.02 &  0.18 $\pm$ 0.02 &   0.02 $\pm$ 0.01 \\
5                      &   0.20 $\pm$ 0.03 &  0.20 $\pm$ 0.03 &   0.00 $\pm$ 0.03 \\
\bottomrule
  \end{tabular}
\end{table}

\begin{table}[ht]
\footnotesize
  \caption{Average test log likelihoods. We report mean plus or minus one standard error over the splits. Bold numbers correspond to the best models overall.}
  \label{tab:dwp:uci_lls}
  \centering
  \begin{tabular}{rcccc}
    \toprule
& & & DWP & \\
\{Dataset\}-\{Depth\} & DGP & $\QGW$ & $\QAGW$ & $\QABGW$ \\
\midrule   
\textsc{\boston} - 2 & -2.43 $\pm$ 0.05 & -2.40 $\pm$ 0.05 & \textbf{-2.37 $\pm$ 0.05} & \textbf{-2.37 $\pm$ 0.05}  \\ 
3 & -2.39 $\pm$ 0.04 & -2.38 $\pm$ 0.05 & \textbf{-2.35 $\pm$ 0.04} & \textbf{-2.35 $\pm$ 0.04} \\ 
4 & -2.41 $\pm$ 0.04 & -2.38 $\pm$ 0.04 & \textbf{-2.37 $\pm$ 0.04} & \textbf{-2.37 $\pm$ 0.04} \\ 
5 & -2.43 $\pm$ 0.04 & -2.38 $\pm$ 0.04 & -2.39 $\pm$ 0.05 & \textbf{-2.38 $\pm$ 0.04} \\ 
\midrule 
\textsc{\concrete} - 2 & -3.10 $\pm$ 0.02 & -3.12 $\pm$ 0.02 & \textbf{-3.08 $\pm$ 0.02} & \textbf{-3.08 $\pm$ 0.02} \\ 
3 & -3.08 $\pm$ 0.02 & -3.10 $\pm$ 0.02 & \textbf{-3.06 $\pm$ 0.02} & -3.07 $\pm$ 0.02 \\ 
4 & -3.13 $\pm$ 0.02 & -3.12 $\pm$ 0.02 & \textbf{-3.07 $\pm$ 0.02} & \textbf{-3.07 $\pm$ 0.02}  \\ 
5 & -3.13 $\pm$ 0.02 & -3.13 $\pm$ 0.02 & \textbf{-3.07 $\pm$ 0.02} & -3.08 $\pm$ 0.02  \\ 
\midrule 
\textsc{\energy} - 2 & -0.70 $\pm$ 0.03 & -0.70 $\pm$ 0.03 & -0.70 $\pm$ 0.03 & -0.70 $\pm$ 0.03  \\ 
3 & -0.70 $\pm$ 0.03 & -0.70 $\pm$ 0.03 & -0.70 $\pm$ 0.03 & -0.70 $\pm$ 0.03  \\ 
4 & -0.70 $\pm$ 0.03 & -0.70 $\pm$ 0.03 & -0.70 $\pm$ 0.03 & -0.70 $\pm$ 0.03 \\ 
5 & -0.71 $\pm$ 0.03 & -0.71 $\pm$ 0.03 & \textbf{-0.70 $\pm$ 0.03} & \textbf{-0.70 $\pm$ 0.03}  \\ 
\midrule 
\textsc{\kinnm} - 2 & 1.35 $\pm$ 0.00 & 1.35 $\pm$ 0.00 & \textbf{1.36 $\pm$ 0.00} & \textbf{1.36 $\pm$ 0.00} \\ 
3 & 1.37 $\pm$ 0.00 & 1.37 $\pm$ 0.00 & \textbf{1.38 $\pm$ 0.00} & \textbf{1.38 $\pm$ 0.00}  \\ 
4 & 1.38 $\pm$ 0.00 & 1.39 $\pm$ 0.01 & \textbf{1.40 $\pm$ 0.00} & \textbf{1.40 $\pm$ 0.00}  \\ 
5 & 1.38 $\pm$ 0.00 & 1.40 $\pm$ 0.01 & \textbf{1.41 $\pm$ 0.01} & \textbf{1.41 $\pm$ 0.01}\\ 
\midrule 
\textsc{\naval} - 2 & \textbf{8.24 $\pm$ 0.06} & 8.23 $\pm$ 0.08 & 8.18 $\pm$ 0.11 & 8.18 $\pm$ 0.13  \\ 
3 & 8.15 $\pm$ 0.06 & 8.18 $\pm$ 0.07 & 8.27 $\pm$ 0.05 & \textbf{8.38 $\pm$ 0.03} \\ 
4 & 8.28 $\pm$ 0.04 & 8.17 $\pm$ 0.11 & 8.14 $\pm$ 0.13 & \textbf{8.32 $\pm$ 0.06} \\ 
5 & 8.28 $\pm$ 0.04 & 8.17 $\pm$ 0.07 & \textbf{8.40 $\pm$ 0.02} & 8.10 $\pm$ 0.19 \\ 
\midrule 
\textsc{\power} - 2 & -2.78 $\pm$ 0.01 & -2.77 $\pm$ 0.01 & \textbf{-2.76 $\pm$ 0.01} & \textbf{-2.76 $\pm$ 0.01} \\ 
3 & -2.77 $\pm$ 0.01 & \textbf{-2.76 $\pm$ 0.01} & \textbf{-2.76 $\pm$ 0.01} & \textbf{-2.76 $\pm$ 0.01}  \\ 
4 & -2.78 $\pm$ 0.01 & -2.77 $\pm$ 0.01 & \textbf{-2.75 $\pm$ 0.01} & \textbf{-2.75 $\pm$ 0.01}  \\ 
5 & -2.78 $\pm$ 0.01 & -2.77 $\pm$ 0.01 & \textbf{-2.76 $\pm$ 0.01} & \textbf{-2.76 $\pm$ 0.01} \\ 
\midrule 
\textsc{\protein} - 2 & -2.82 $\pm$ 0.00 & \textbf{-2.81 $\pm$ 0.00} & \textbf{-2.81 $\pm$ 0.00} & \textbf{-2.81 $\pm$ 0.00} \\ 
3 & -2.78 $\pm$ 0.00 & -2.77 $\pm$ 0.00 & \textbf{-2.76 $\pm$ 0.00} & \textbf{-2.76 $\pm$ 0.00}  \\ 
4 & -2.75 $\pm$ 0.00 & -2.73 $\pm$ 0.00 & \textbf{-2.72 $\pm$ 0.00} & -2.73 $\pm$ 0.01  \\ 
5 & -2.73 $\pm$ 0.01 & -2.72 $\pm$ 0.01 & -2.71 $\pm$ 0.01 & \textbf{-2.70 $\pm$ 0.00} \\ 
\midrule 
\textsc{\wine} - 2 & -0.96 $\pm$ 0.01 & -0.96 $\pm$ 0.01 & -0.96 $\pm$ 0.01 & -0.96 $\pm$ 0.01 \\ 
3 & -0.96 $\pm$ 0.01 & -0.96 $\pm$ 0.01 & -0.96 $\pm$ 0.01 & -0.96 $\pm$ 0.01 \\ 
4 & -0.96 $\pm$ 0.01 & -0.96 $\pm$ 0.01 & -0.96 $\pm$ 0.01 & -0.96 $\pm$ 0.01\\ 
5 & -0.96 $\pm$ 0.01 & -0.96 $\pm$ 0.01 & -0.96 $\pm$ 0.01 & -0.96 $\pm$ 0.01  \\ 
\midrule 
\textsc{\yacht} - 2 & -0.29 $\pm$ 0.12 & \textbf{-0.04 $\pm$ 0.10} & \textbf{-0.04 $\pm$ 0.08} & -0.08 $\pm$ 0.10  \\ 
3 & -0.63 $\pm$ 0.04 & -0.13 $\pm$ 0.07 & 0.12 $\pm$ 0.07 & \textbf{0.14 $\pm$ 0.06} \\ 
4 & -0.77 $\pm$ 0.07 & -0.26 $\pm$ 0.07 & \textbf{-0.04 $\pm$ 0.09} & \textbf{-0.04 $\pm$ 0.09}  \\ 
5 & -0.73 $\pm$ 0.07 & -0.58 $\pm$ 0.06 & -0.22 $\pm$ 0.09 & \textbf{-0.18 $\pm$ 0.07} \\ 
\bottomrule
  \end{tabular}
\end{table}

\begin{table}[ht]
\footnotesize
  \caption{Root mean square error. We report mean plus or minus one standard error over the splits. Bold numbers correspond to the best models overall.}
  \label{tab:dwp:uci_rmses}
  \centering
  \begin{tabular}{rcccc}
    \toprule
& & & DWP & \\
\{Dataset\}-\{Depth\} & DGP & $\QGW$ & $\QAGW$ & $\QABGW$ \\
\midrule  
\textsc{\boston} - 2 & 2.72 $\pm$ 0.14 & 2.67 $\pm$ 0.14 & 2.60 $\pm$ 0.12 & \textbf{2.59 $\pm$ 0.13} \\ 
3 & 2.73 $\pm$ 0.14 & 2.66 $\pm$ 0.13 & \textbf{2.62 $\pm$ 0.13} & 2.63 $\pm$ 0.13 \\ 
4 & 2.76 $\pm$ 0.14 & 2.74 $\pm$ 0.15 & 2.71 $\pm$ 0.14 & \textbf{2.68 $\pm$ 0.14} \\ 
5 & 2.81 $\pm$ 0.14 & 2.82 $\pm$ 0.17 & \textbf{2.77 $\pm$ 0.16} & 2.81 $\pm$ 0.17 \\ 
\midrule 
\textsc{\concrete} - 2 & 5.41 $\pm$ 0.10 & 5.50 $\pm$ 0.12 & \textbf{5.29 $\pm$ 0.12} & 5.30 $\pm$ 0.12 \\ 
3 & 5.31 $\pm$ 0.11 & 5.32 $\pm$ 0.10 & \textbf{5.22 $\pm$ 0.12} & 5.23 $\pm$ 0.12 \\ 
4 & 5.54 $\pm$ 0.10 & 5.43 $\pm$ 0.11 & 5.24 $\pm$ 0.13 & \textbf{5.22 $\pm$ 0.13} \\ 
5 & 5.49 $\pm$ 0.10 & 5.53 $\pm$ 0.10 & 5.26 $\pm$ 0.11 & \textbf{5.24 $\pm$ 0.11} \\ 
\midrule 
\textsc{\energy} - 2 & 0.48 $\pm$ 0.01 & 0.48 $\pm$ 0.01 & 0.48 $\pm$ 0.01 & 0.48 $\pm$ 0.01 \\ 
3 & 0.48 $\pm$ 0.01 & 0.48 $\pm$ 0.01 & 0.48 $\pm$ 0.01 & 0.48 $\pm$ 0.01 \\ 
4 & 0.48 $\pm$ 0.01 & 0.48 $\pm$ 0.01 & 0.48 $\pm$ 0.01 & 0.48 $\pm$ 0.01 \\ 
5 & 0.49 $\pm$ 0.01 & \textbf{0.48 $\pm$ 0.01} & \textbf{0.48 $\pm$ 0.01} & \textbf{0.48 $\pm$ 0.01} \\ 
\midrule 
\textsc{\kinnm} - 2 & 0.06 $\pm$ 0.01 & 0.06 $\pm$ 0.01 & 0.06 $\pm$ 0.00 & 0.06 $\pm$ 0.00 \\ 
3 & 0.06 $\pm$ 0.01 & 0.06 $\pm$ 0.01 & 0.06 $\pm$ 0.00 & 0.06 $\pm$ 0.00 \\ 
4 & 0.06 $\pm$ 0.01 & 0.06 $\pm$ 0.01 & 0.06 $\pm$ 0.00 & 0.06 $\pm$ 0.00 \\ 
5 & 0.06 $\pm$ 0.01 & 0.06 $\pm$ 0.01 & 0.06 $\pm$ 0.00 & 0.06 $\pm$ 0.00 \\ 
\midrule 
\textsc{\naval} - 2 & 0.00 $\pm$ 0.00 & 0.00 $\pm$ 0.00 & 0.00 $\pm$ 0.00 & 0.00 $\pm$ 0.00 \\ 
3 & 0.00 $\pm$ 0.00 & 0.00 $\pm$ 0.00 & 0.00 $\pm$ 0.00 & 0.00 $\pm$ 0.00 \\ 
4 & 0.00 $\pm$ 0.00 & 0.00 $\pm$ 0.00 & 0.00 $\pm$ 0.00 & 0.00 $\pm$ 0.00 \\ 
5 & 0.00 $\pm$ 0.00 & 0.00 $\pm$ 0.00 & 0.00 $\pm$ 0.00 & 0.00 $\pm$ 0.00 \\ 
\midrule 
\textsc{\power} - 2 & 3.87 $\pm$ 0.04 & 3.83 $\pm$ 0.04 & 3.82 $\pm$ 0.04 & \textbf{3.81 $\pm$ 0.04} \\ 
3 & 3.87 $\pm$ 0.03 & 3.82 $\pm$ 0.04 & \textbf{3.81 $\pm$ 0.04} & \textbf{3.81 $\pm$ 0.04} \\ 
4 & 3.89 $\pm$ 0.04 & 3.84 $\pm$ 0.04 & \textbf{3.78 $\pm$ 0.04} & \textbf{3.78 $\pm$ 0.04} \\ 
5 & 3.88 $\pm$ 0.04 & 3.84 $\pm$ 0.04 & \textbf{3.80 $\pm$ 0.04} & \textbf{3.80 $\pm$ 0.04} \\ 
\midrule 
\textsc{\protein} - 2 & 4.08 $\pm$ 0.01 & 4.06 $\pm$ 0.01 & \textbf{4.05 $\pm$ 0.02} & \textbf{4.05 $\pm$ 0.01} \\ 
3 & 3.92 $\pm$ 0.02 & 3.90 $\pm$ 0.01 & 3.88 $\pm$ 0.01 & \textbf{3.87 $\pm$ 0.01} \\ 
4 & 3.82 $\pm$ 0.01 & 3.79 $\pm$ 0.01 & \textbf{3.75 $\pm$ 0.01} & 3.79 $\pm$ 0.02 \\ 
5 & 3.77 $\pm$ 0.02 & 3.76 $\pm$ 0.02 & 3.73 $\pm$ 0.02 & \textbf{3.70 $\pm$ 0.01} \\ 
\midrule 
\textsc{\wine} - 2 & 0.63 $\pm$ 0.01 & 0.63 $\pm$ 0.01 & 0.63 $\pm$ 0.01 & 0.63 $\pm$ 0.01 \\ 
3 & 0.63 $\pm$ 0.01 & 0.63 $\pm$ 0.01 & 0.63 $\pm$ 0.01 & 0.63 $\pm$ 0.01 \\ 
4 & 0.63 $\pm$ 0.01 & 0.63 $\pm$ 0.01 & 0.63 $\pm$ 0.01 & 0.63 $\pm$ 0.01 \\ 
5 & 0.63 $\pm$ 0.01 & 0.63 $\pm$ 0.01 & 0.63 $\pm$ 0.01 & 0.63 $\pm$ 0.01 \\ 
\midrule 
\textsc{\yacht} - 2 & 0.41 $\pm$ 0.04 & \textbf{0.33 $\pm$ 0.03} & \textbf{0.33 $\pm$ 0.03} & \textbf{0.33 $\pm$ 0.03} \\ 
3 & 0.53 $\pm$ 0.03 & 0.35 $\pm$ 0.03 & 0.31 $\pm$ 0.03 & \textbf{0.30 $\pm$ 0.03} \\ 
4 & 0.58 $\pm$ 0.05 & 0.41 $\pm$ 0.04 & \textbf{0.33 $\pm$ 0.03} & \textbf{0.33 $\pm$ 0.03} \\ 
5 & 0.57 $\pm$ 0.05 & 0.50 $\pm$ 0.04 & \textbf{0.37 $\pm$ 0.03} & 0.38 $\pm$ 0.03 \\ 
\bottomrule
  \end{tabular}
\end{table}
  }{}

\end{document}